\documentclass[10pt,twocolumn,letterpaper]{article}

\usepackage[pagenumbers]{cvpr} 

\usepackage[dvipsnames]{xcolor}

\definecolor{cvprblue}{rgb}{0.21,0.49,0.74}
\usepackage[pagebackref,breaklinks,colorlinks,citecolor=cvprblue]{hyperref}

\usepackage{algorithm}
\usepackage{algorithmic}

\usepackage{booktabs} 
\usepackage{colortbl}
\usepackage{pifont}

\usepackage{tabularx}
\usepackage{multirow}
\usepackage{subcaption}
\DeclareCaptionLabelFormat{subfig}{#1~#2}
\usepackage{adjustbox}
\usepackage{transparent}

\usepackage{epsfig}
\usepackage{amsmath}
\usepackage{amssymb}
\usepackage{float}
\usepackage{lipsum}

\usepackage{soul}
\usepackage{setspace}
\usepackage{marvosym}

\usepackage{xcolor}

\definecolor{sotacolor}{RGB}{220, 239, 220}
\def\paperID{6594} 
\def\confName{CVPR}
\def\confYear{2024}

\title{Gaussian-Flow: 4D Reconstruction with Dynamic 3D Gaussian Particle}

\begin{spacing}{3}
\author{Youtian Lin$^1$ \qquad Zuozhuo Dai$^2$ \qquad Siyu Zhu$^3$ \qquad Yao Yao$^1$\textsuperscript{\Letter}\\
$^{1}$Nanjing University \qquad $^{2}$Alibaba Group \qquad $^{3}$Fudan University} 
\end{spacing}

\begin{document}
\twocolumn[{%
\renewcommand\twocolumn[1][]{#1}%
\maketitle
\begin{center}
    \centering
    \captionsetup{type=figure}
    \includegraphics[width=1.\linewidth]{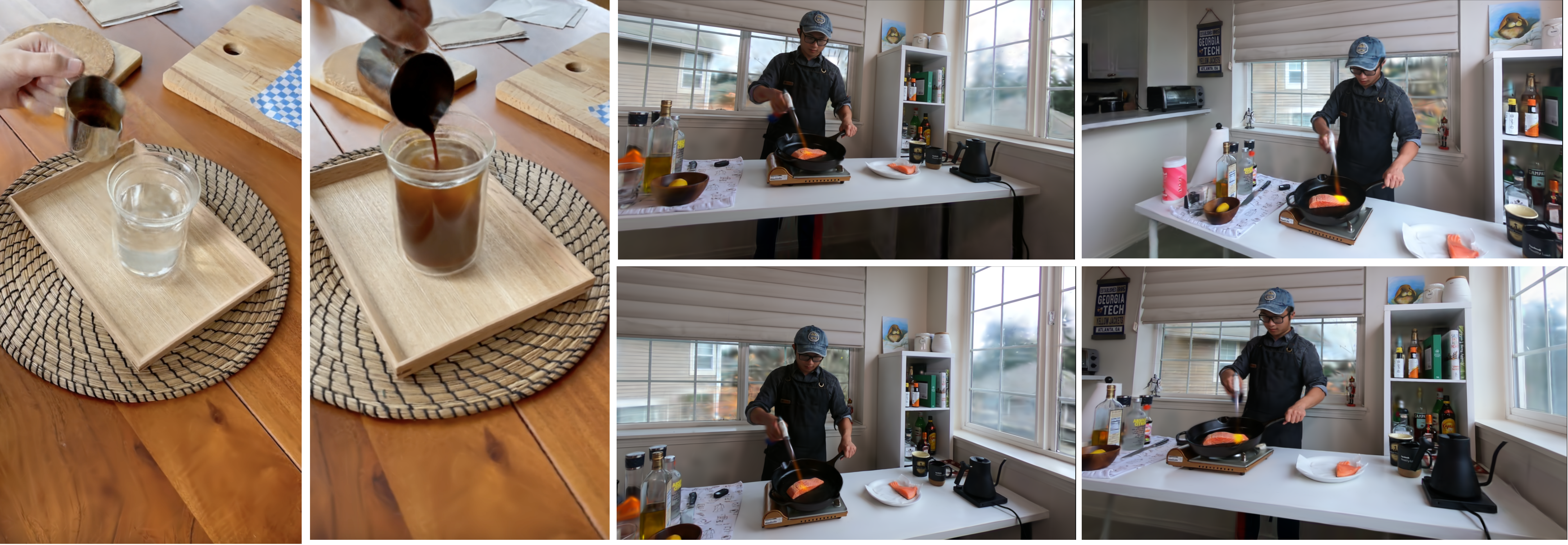}
    \captionof{figure}{Dynamic reconstruction results of the proposed Gaussian-Flow on the monocular HyperNeRF Dataset~\cite{park2021hypernerf} (left) and the multi-view Plenoptic Dataset~\cite{Li_2022_CVPR} (right). Our method achieves a $5\times$ faster training and rendering speed compared with the per-frame 3DGS modeling and significantly outperforms previous methods in novel view rendering quality.}
    \label{fig:teaser}
\end{center}%
}]

\begin{abstract}

We introduce Gaussian-Flow, a novel point-based approach for fast dynamic scene reconstruction and real-time rendering from both multi-view and monocular videos. In contrast to the prevalent NeRF-based approaches hampered by slow training and rendering speeds, our approach harnesses recent advancements in point-based 3D Gaussian Splatting (3DGS). Specifically, a novel Dual-Domain Deformation Model (DDDM) is proposed to explicitly model attribute deformations of each Gaussian point, where the time-dependent residual of each attribute is captured by a polynomial fitting in the time domain, and a Fourier series fitting in the frequency domain. The proposed DDDM is capable of modeling complex scene deformations across long video footage, eliminating the need for training separate 3DGS for each frame or introducing an additional implicit neural field to model 3D dynamics. 
Moreover, the explicit deformation modeling for discretized Gaussian points ensures ultra-fast training and rendering of a 4D scene, which is comparable to the original 3DGS designed for static 3D reconstruction. Our proposed approach showcases a substantial efficiency improvement, achieving a $5\times$ faster training speed compared to the per-frame 3DGS modeling. In addition, quantitative results demonstrate that the proposed Gaussian-Flow significantly outperforms previous leading methods in novel view rendering quality. Project page: \url{https://nju-3dv.github.io/projects/Gaussian-Flow}.

\end{abstract}    
\section{Introduction}
\label{sec:intro}

In the realm of digital scene synthesis, achieving a balance between high-quality reconstructions and real-time rendering is paramount, especially for applications like virtual reality (VR) playback, where immediate feedback and immersive experiences are essential. Neural Radiance Fields (NeRFs)~\cite{mildenhall2020nerf} has risen as a promising method for synthesizing intricate scenes. However, despite their ability to produce visually stunning results, NeRFs require costly sampling and evaluation of the neural radiance field at multiple points along each ray. Consequently, the substantial computational demands impede the real-time rendering capabilities of NeRFs.
These are attempts to accelerate the rendering process of NeRFs, such as direct volume representation~\cite{nsvf,PlenOctrees,dvgo}, neural hasing~\cite{instantngp}, and tri-plane structures~\cite{tensorf,HexPlane,kplanes}.
However, it still remains a challenge for high-fidelity real-time rendering. Nevertheless, the issue becomes even more severe when turning to dynamic scene reconstruction and rendering.

Recent progress on 3D Gaussian Splatting (3DGS)~\cite{kerbl3Dgaussians} has drawn attention from the 3D computer vision community. With tile-based rasterization instead of plain volume rendering, 3DGS can render images two orders of magnitude faster than the vanilla NeRF. The technique has also been quickly applied to 4D scene reconstruction by extending to the separate per-frame 3DGS optimization~\cite{luiten2023dynamic}. However, such direct extension is storage-intensive and is not applicable to monocular video input. Some other concurrent works~\cite{wu20234d,yang2023deformable3dgs} try to mix the explicit point-based 3DGS and an implicit neural field for dynamic information modeling, however, require computationally expensive forward passes of the neural network, which significantly lowers the rendering speed of the original 3DGS. 

In this work, we propose Gaussian-Flow, an explicit particle-based deformation model designed specifically for 3DGS to model the dynamic scene without using any neural network. Gaussian-Flow can recover a high-fidelity 4D scene from captured videos, while still preserving the ultra-fast training and rendering speed of the original 3DGS. In particular, we formulate a 4D scene as a set of deformable 3D Gaussian points. A novel Dual-Domain Deformation Model (DDDM) is proposed to explicitly model deformations of each Gaussian point's attributes, including position, rotation, and radiance. The time-dependent deformation residual is modeled simultaneously in time and frequency domains: we apply joint polynomial and Fourier series fitting for each deformable attribute. This compact dynamic representation greatly reduces the computation cost of the deformation model, which is a key factor in preserving the rendering speed of 3DGS. Moreover, an adaptive timestamp scaling technique is introduced to avoid over-fitting the scene to only frames with violent motions. For robust estimation, we also regularize the motion trajectory by the KNN-based rigid and the time smooth constraints. It is also noteworthy that our discretized point-based 4D representation naturally supports the edition of both static and dynamic 3D scenes, showing the potential for unlocking a variety of downstream applications related to dynamic 3D reconstruction and rendering.  

We have conducted extensive experiments to demonstrate the effectiveness of the proposed method on several multi-view and monocular datasets. The proposed Gaussian-Flow achieves a $5\times$ faster training speed compared with the separate per-frame 3DGS modeling, and significantly outperforms prior leading methods in novel view rendering quality. Our major contributions can be summarized as follows:
\begin{itemize}
    \item We introduce Gaussian-Flow, a novel point-based differentiable rendering approach for dynamic 3D scene reconstruction, setting a new state-of-the-art for training speed, rendering FPS, and novel view synthesis quality for 4D scene reconstruction.
    \item We propose a Dual-Domain Deformation Model for efficient 4D scene training and rendering, eliminating the need for per-frame 3DGS optimization and sampling on implicit neural fields. This preserves a running speed on par with the original 3DGS with minimum overhead.
    \item We demonstrate that our discretized point-based representation supports the segmentation, edition, and composition of both static and dynamic 3D scenes.  
\end{itemize}

\section{Related Works}
\label{sec:relate_works}

\subsection{Dynamic Neural Radiance Field}

Dynamic NeRF modeling has become a heated research topic in recent years due to the development of the neural radiance field and differentiable rendering. 
By treating time as an extended input dimension to NeRF, researchers successfully achieve qualified image-based 4D scene rendering~\cite{nsff,xian2020space,gao2021dynamic,du2021neural, lombardi2019neural}. 
To further improve the reconstruction quality and incorporate prior knowledge of motions and structures, dynamic neural scene flow methods have been proposed~\cite{pumarola2020dnerf,nerfies}, where a canonical space is constructed and then transferred to each frame with scene flow or motion fields. 
HyperNeRF~\cite{park2021hypernerf} models the deformation of the object topologies by using higher-dimensional inputs, while DyNeRF~\cite{Li_2022_CVPR} utilizes time-conditioned NeRF to represent a 4D scene. However, the aforementioned approaches are all based on the vanilla NeRF, which requires a long training time and does not meet the requirement of real-time rendering. 


\subsection{Accelerated Neural Radiance Field}

To expedite NeRF training and rendering, numerous approaches have been suggested, employing more streamlined strategies~\cite{fang2021neusample, autoint, Piala2021TermiNeRFRT, Garbin_2021_ICCV, Hedman_2021_ICCV, wu2021diver, Kurz2022AdaNeRFAS}.
Other methods propose the integration of neural implicit functions with explicit 3D structures, forming a hybrid representation for faster radiance field sampling~\cite{instantngp, dvgov2, dvgo, tensorf}. These approaches establish strong foundations for enhancing dynamic NeRF, concurrently decreasing required training and inference time. Apart from implicit neural representations, explicit NeRF modeling has show promising results for real-time rendering: NSVF~\cite{nsvf} employs a neural sparse voxel field for efficient NeRF sampling, which stands for the earliest attempt on the explicit NeRF modeling; PlenOctrees~\cite{PlenOctrees} utilizes the explicit octree structure for rendering acceleration. 

Recent efforts have also emerged to accelerate the intricate dynamic neural radiance field. TensorRF~\cite{tensorf} employs multiple planes as explicit representations for direct dynamic scene modeling. More recent recent approaches of this kind include K-Planes~\cite{fridovich2023k}, Tensor4D~\cite{shao2022tensor4d}, and HexPlane~\cite{cao2023hexplane}. Alternatively, NeRFPlayer\cite{song2022nerfplayer} introduces a unified streaming representation for both grid-based~\cite{instantngp} and plane-based methods, utilizing separate models to distinguish static and dynamic scene components, however, leading to slow rendering times. HyperReel~\cite{attal2023hyperreel} further suggests a flexible sampling network coupled with two planes for dynamic scene representation. While these methods improve the rendering speed of a dynamic scene to some extent, it is still hard to achieve real-time rendering, let alone a good balance between the running speed and rendering quality. In contrast, we resort to the recent 3D Gaussians splatting, which applies an explicit soft point cloud representation for real-time image-based rendering.

\subsection{Differentiable Point-based Rendering}

The original idea of using 3D points as rendering primitives was first introduced in~\cite{levoy1985use}. By incorporating differentiable rendering, recent approaches have made remarkable progress in image-based rendering, representative methods include PointRF~\cite{zhang2022differentiable}, DSS~\cite{yifan2019differentiable}, and 3D Gaussians splatting (3DGS)~\cite{kerbl20233d}. Specifically, 3DGS~\cite{kerbl20233d} has demonstrated extraordinary performance in novel-view synthesis, achieving real-time rendering speed and state-of-the-art rendering quality. The method adopts a soft point representation with attributes of position, rotation, density, and radiance, and applies differentiable point-based rendering for scene optimization. 3DGS has quickly been extended to dynamic scene modeling by direct separate per-frame optimization~\cite{luiten2023dynamic}, however, requires a long optimization time and a large amount of storage for long video footage. Other works~\cite{wu20234d,yang2023deformable3dgs} apply an implicit motion field to model scene dynamics, but the introduction of the implicit neural network significantly slows down the sampling and rendering speed. In this work, we adopt the approach of representing 4D scenes through a purely discretized point cloud model, ensuring a fast training and rendering speed comparable with the original 3DGS.

\section{Gaussian-Flow}
\label{sec:method}

In this section, we introduce the proposed Gaussian-Flow for dynamic scene modeling. We first review the 3DGS in Sec.~\ref{sec:3d-splats}. Then, we introduce our explicit motion modeling of each Gaussian point by using a novel Dual-Domain Deformation Model (DDDM), as outlined in Sec.~\ref{sec:fourier-series}. An adaptive timestamp scaling technique is described in Sec.~\ref{sec:ats} for balanced training of each frame. To ensure the continuity of the motion in both spatial and temporal dimensions, we incorporate appropriate regularizations on each point during the optimization, as detailed in Sec.~\ref{sec:trajectory-motion-regularization}. 

\subsection{Recap on 3D Gaussian Splatting}
\label{sec:3d-splats}
3D Gaussian Splatting~\cite{kerbl3Dgaussians} is designed to efficiently optimize a 3D scene for real-time and high-quality novel view synthesis. The 3DGS framework has garnered significant attention within the community due to its remarkable enhancements in both training and rendering times, concurrently achieving state-of-the-art rendering quality. In contrast to the volume rendering in the vanilla NeRF which relies on ray marching, 3DGS adopts a tile-based rasterization on a distinctive soft point cloud representation to achieve fast rendering. Specifically, 3DGS models a 3D scene as a large amount of 3D Gaussian points in the world space, where each point is represented by:
\begin{equation}
\label{eq:3d_gaussian}
G(\boldsymbol{x}) = \exp(-\frac{1}{2}(\boldsymbol{x}-\boldsymbol{\mu})^T\boldsymbol{\Sigma}^{-1}(\boldsymbol{x}-\boldsymbol{\mu})),
\end{equation}
where $\boldsymbol{\mu}$ and $\boldsymbol{\Sigma}$ are the mean position and covariance matrix of a 3D Gaussian particle. The 3DGS takes sparse Structure-from-Motion (SfM) points or even random points as input, and initializes each point to a 3D Gaussian based on its neighbors. Besides, each 3D Gaussian is associated with a learnable view-dependent radiance $\boldsymbol{c}$ and a learnable opacity $\alpha$ for rendering. Subsequently, an efficient 3D to 2D Gaussian mapping~\cite{zwicker2001ewa} is employed to project the point onto the image plane:
\begin{equation}
\label{eq:projection1}
\boldsymbol{\mu}' = \boldsymbol{P}\boldsymbol{W}{\boldsymbol{{\mu}}}, 
\end{equation}
\begin{equation}
\label{eq:projection2}
\boldsymbol{\Sigma}' = \boldsymbol{J} \boldsymbol{W} \boldsymbol{\Sigma} \boldsymbol{W}^T \boldsymbol{J}^T,
\end{equation}
where $\boldsymbol{\mu}'$ and $\boldsymbol{\Sigma}'$ represent the 2D mean position and 2D covariance of the projected 3D Gaussian. $\boldsymbol{P}$,$\boldsymbol{W}$ and $\boldsymbol{J}$ donate the projective transformation, viewing transformation, and Jacobian of the affine approximation of $\boldsymbol{P}$. After that, $\alpha$-blending is executed to merge the overlapping Gaussians for each pixel, yielding a final color:
\begin{equation}
    \label{eq:3dgs-rasterization}
    \begin{aligned}
        C &= \sum_{i=1}^{n} \boldsymbol{c}_i \alpha_i \prod_{j=1}^{i-1} (1 - \alpha_i),
    \end{aligned}
\end{equation}
where $\boldsymbol{c}_i$ and $\alpha_i$ are the color and opacity of the $i$-th Gaussian, and $n$ is the number of overlapping Gaussians.

The attributes of the 3D Gaussian, including the mean $\boldsymbol{\mu}$, covariance matrix $\boldsymbol{\Sigma}$, opacity $\alpha$, and color $\boldsymbol{c}$, are optimized through backward propagation of the gradient flow. In particular, to ensure positive definiteness during the optimization process, the covariance matrix is parameterized as a scaling vector $\boldsymbol{s}$ and a rotation matrix $\boldsymbol{R}$, i.e., $\boldsymbol{\Sigma} = \boldsymbol{R}\Lambda(\boldsymbol{s})\Lambda(\boldsymbol{s})^T\boldsymbol{R}^T$, where $\Lambda(\boldsymbol{s})$ is the diagonal matrix of $\boldsymbol{s}$. To facilitate optimization, the rotation matrix is further parameterized using a quaternion $\boldsymbol{q}$. Leveraging the inherent flexibility of discrete points, 3DGS incorporates adaptive density control for points. This mechanism utilizes the gradient flow to identify where geometric reconstruction is suboptimal, and employs cloning and splitting to augment the density of points for a higher rendering quality.

\begin{figure*}[h]
    \centering
    \includegraphics[width=1.0\textwidth]{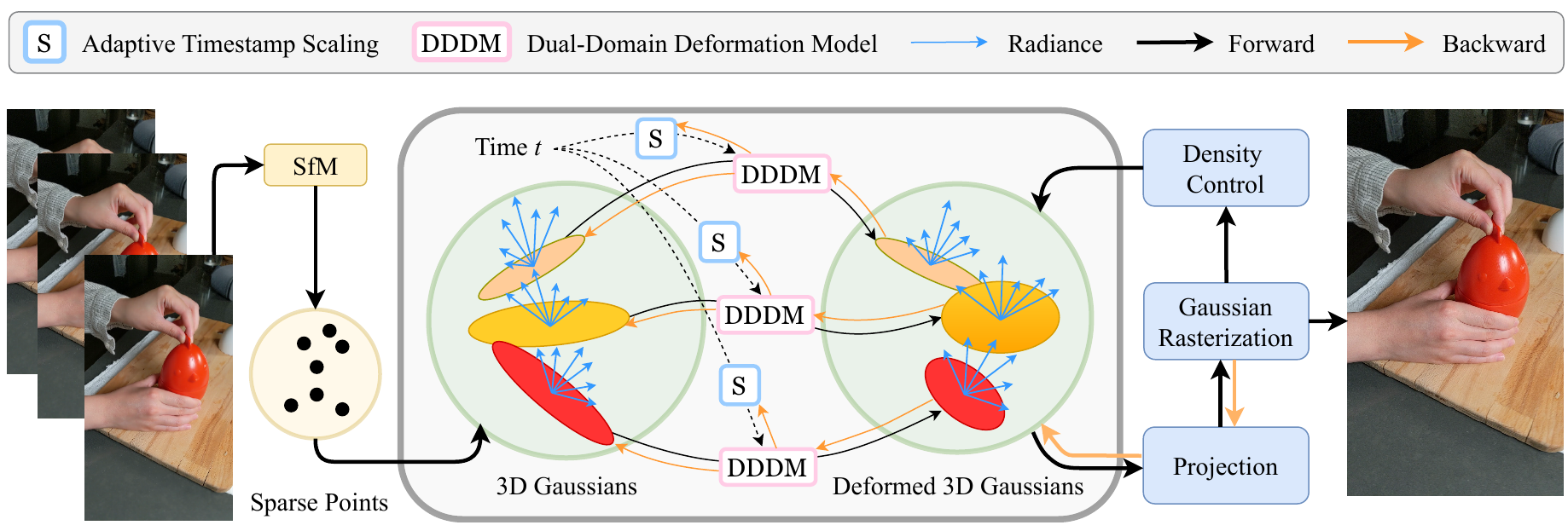}
    \caption{\textbf{Overview of the Gaussian-Flow pipeline}. We model the deformation of attributes of each 3D Gaussian point independently by using the Dual-Domain Deformation Model (DDDM), which preserves the discretized nature of the 3D Gaussian points, and thus achieves ultra-fast training and rendering speed comparable with the original 3DGS.}
    \label{fig:overview}
  \end{figure*}

\subsection{Dual-Domain Deformation Model}
\label{sec:fourier-series}
We target to directly model the dynamics of each 3D Gaussian point by fitting each of its attributes into a time-dependent curve. Among different approaches, Polynomials fitting in the time domain and Fourier series fitting in the frequency domain are the two most widely used approaches~\cite{nsff,xian2020space,gao2021dynamic,du2021neural,lombardi2019neural}, due to their simplicity and effectiveness. However, each method comes with its own advantages and drawbacks: describing the motion of a Gaussian particle in terms of polynomials yields a good fit with smooth motion with a small order of polynomials, however, can easily overfit to a violent motion if assuming a larger order of polynomials, resulting in unreasonable oscillations in the fitted trajectory. Whereas, the Fourier series excels at capturing the variations associated with violent motion, however, requires a manually reduced order when dealing with smooth motion.

\begin{figure}[ht]
    \centering
    \begin{subfigure}[b]{0.45\linewidth}
        \includegraphics[width=\linewidth]{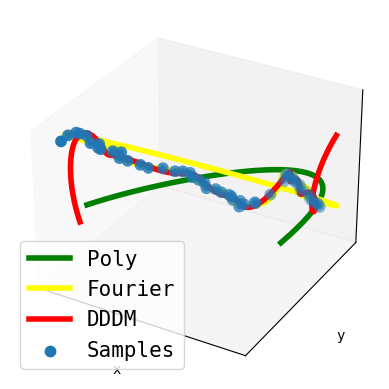}
    \end{subfigure}
    \begin{subfigure}[b]{0.45\linewidth}
        \includegraphics[width=\linewidth]{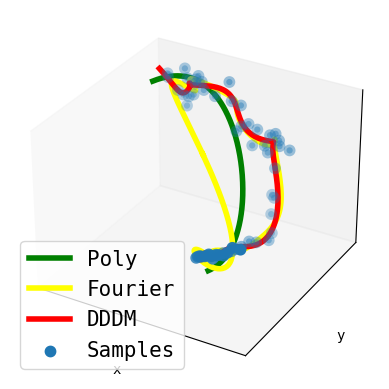}
    \end{subfigure}
    \caption{Two exemplar motion fittings using polynomial, Fourier Series, and the joint DDDM functions. Our DDDM is able to accurately fit complex trajectories denoted by the sampling points.}
    \label{fig:fft-vs-poly-vs-dddm}
    \end{figure}

In this work, our key insight is to use a Dual-Domain Deformation Model (DDDM) for fitting the scene dynamics, which integrates both the time domain polynomials and the frequency domain Fourier series into a unified fitting model. We assume that only the rotation $\boldsymbol{q}$, radiance $\boldsymbol{c}$, and position $\boldsymbol{\mu}$ of a 3D Gaussian particle change over time, while the scaling $\boldsymbol{s}$ and opacity $\alpha$ remain constant. Specifically, we conceptualize the change in each particle's attributes as its base attributes $\boldsymbol{S}_0\in\{\boldsymbol{\mu}_0, \boldsymbol{c}_0, \boldsymbol{q}_0\}$ at the reference time frame $t_0$ (usually set to the first frame), superimposed on a time-dependent attribute residual $\boldsymbol{D}(t)$. For simplicity, we use lowercase characters to represent a single attribute in $\boldsymbol{S}$. The time-dependent residual of each attribute is modeled through polynomial fitting in the time domain and Fourier series fitting in the frequency domain, expressed as:
\begin{equation}
    \label{eq:residual}
    S(t) = S_0 + D(t),
\end{equation}
where $D(t)=P_N(t)+F_L(t)$ is combined by a polynomial $P_N(t)$ with coefficients $\boldsymbol{a}=\{a\}^{N}_{n=0}$ and a Fourier series $F_L(t)$ with coefficients $\boldsymbol{f} = \{f^{l}_{sin},f^{l}_{cos}\}^{L}_{l=0}$. These are respectively denoted as:
\begin{equation}
    \label{eq:polynomial}
    P_N(t) = \sum_{n=0}^{N}{a_n}{t^n},
\end{equation}
\begin{equation}
    \label{eq:fourier-series}
    F_L(t) = \sum_{l=1}^{L} \left( f^l_{sin} \cos(lt) + f^l_{cos} \sin(lt) \right).
\end{equation}
It is important to note that we assume different dimensions of an attribute are independently changed over time. Therefore, we assign a different $D(t)$ for each dimension of an attribute. For instance, we utilize $\{D_{\mu_i}(t)\}_{i=0}^3$ to describe the motion of a 3D position $\boldsymbol{\mu}$.

Figure~\ref{fig:fft-vs-poly-vs-dddm} illustrates a comparative analysis of trajectory fitting using polynomial, Fourier series, and the proposed joint DDDM functions. The figure highlights the superior fitting capabilities of the DDDM approach in capturing complex motion trajectories as represented by the sampled data points.

\subsection{Adaptive Timestemp Scaling}
\label{sec:ats}
In a typical scenario, a normalized frame index $t$ ranging from 0 to 1 will be used as the temporal input of $D(t)$. However, this poses a challenge when endeavoring to model substantial motions within a very short time using polynomials and Fourier series. Adhering to the standard temporal division would necessitate an exceedingly large coefficient to accommodate highly intense movements within a very short time frame. This circumstance has the potential to induce instability or even a breakdown in the optimization process. To address this issue, we introduce a time dilation factor $\lambda$ to scale the temporal input for each Gaussian point, which is formulated as:
\begin{equation}
     t_s = \lambda_s \cdot t + \lambda_b
\end{equation}
where $t_s$ represents the scaled time input, serving as the input of $D(t)$, $t \in [0, 1]$ denotes the normalized frame index, and $\lambda$ and $\lambda_b$ stand for the dilation factor and base factor of a Gaussian, respectively. In all our experiments, $\lambda_s$ and $\lambda_b$ are initialized to 1 and 0, respectively.

To summarize, in our dynamic scene setting, a Gaussian particle contains multiple attributes to be optimized, including base attributes $\{\mu_0, q_0, s_0, c_0, \alpha_0\}$ at reference frame $t_0$, polynomial coefficients and Fourier coefficients in $\{D_{\boldsymbol{\mu}}(t), D_{\boldsymbol{q}}(t), D_{\boldsymbol{c}}(t)\}$. Since the proposed DDDM optimizes the time-dependent residuals without the need for an intricate neural field structure, our Gaussian-Flow inherits the benefits of extreme-fast training and rendering speed from the vanilla 3DGS.

\subsection{Regularizations}
\label{sec:trajectory-motion-regularization}
While the utilization of discrete points as scene representations accelerates rendering, several challenges remain. First, points are optimized individually, losing connections with their spatial neighbors, which do not align with the real-world scenario. Optimizing these Gaussian points without considering continuity will inevitably lead to a degradation in reconstruction quality and spatial coherence. Additionally, motions should be smoothed over time. Based on these observations, we employ two regularizations, namely time smoothness and a KNN rigid regularization, for robust optimization of Gaussian points and their motions.

\paragraph{Time Smoothness Loss}
To ensure temporal smoothness over time, we apply a perturbing $\epsilon$ on the input timestamp $t$ and encourage the time-dependent attributes (i.e., position $\mu$, rotation $q$ and radiance $c$) at time $t+\epsilon$ to be consistent with those at time $t$. The time smoothness term is defined as:
\begin{equation}
    \label{eq:time_smooth}
    \mathcal{L}_t = \Vert D(t) - D(t+\epsilon) \Vert_{2}.
\end{equation}
It is noteworthy that the magnitude of the perturbing value is adaptively set according to the number of total frames, i.e., $\epsilon = 0.1/\text{frames}$.


\paragraph{KNN Rigid Loss}
During the optimization of 3D Gaussians with adaptive density control, points will be dynamically added or removed. This dynamic nature implies that the neighbors of points within a local space are subject to constant changes, posing a challenge for directly enforcing a spatial local consistency constraint. To sidestep this problem, we propose to divide the optimization part into two alternating stages: in the former stage, we optimize all variables with adaptive density control; while in the latter stage, we optimize the attributes without adding or removing points. The local rigid constraint is incorporated in every latter stage, and it is defined as:
\begin{equation}
    \label{eq:time_smooth}
    \mathcal{L}_s = \sum_{j \in \mathcal{N}_i} \Vert D(t)_i - D(t)_j \Vert_{2}
\end{equation}
where, $\mathcal{N}_i$ represents the $K$ nearst neighbor (KNN) of $i$-th Gaussian.

\section{Experiments}

\subsection{Implementation Details}

We train our model using the Adam~\cite{adam} optimizer with separate learning rates for different attributes of the Gaussian point. We set a learning rate of  $4\times10^{-4}$ for the point position with an exponential decay of $8\times10^{-7}$. The learning rates for point rotation and all DDDM parameters are set to $0.002$ and $4\times10^{-4}$ respectively. We apply a weight decay of $8\times10^{-7}$ to all parameters. The rest of the learning rate follows the 3DGS setting. We train the model for 30K steps and 60K steps for all scenes. All experiments are conducted on a single NVIDIA RTX 4090 GPU with 24GB memory. In addition, we use Taichi to implement our DDDM model, which can parallelize the computation of the DDDM (i.e. polynomial and Fourier series computation) of each Gaussian point.

\subsection{Datasets}

We evaluate our method on both multi-view and monocular datasets, to demonstrate the effectiveness of our method in both settings. 

\paragraph{Plenoptic Video dataset~\cite{Li_2022_CVPR}}
The dataset was captured using 21 cameras at a resolution of 2704 $\times$ 2028, with each camera recording a 10-second video. Six scenes from this dataset are publicly available. For a fair comparison, we downsampled the images to 1352 $\times$ 1014 resolution in our experiments to keep the same setting from the concurrent work of 4D Gaussian~\cite{wu20234d}.

\paragraph{HyperNeRF dataset~\cite{park2021hypernerf}}
This dataset uses a monocular camera (e.g., iPhone) to record real-world motions, which includes real rigid and non-rigidly deforming scenes, such as a person splitting a cookie. The dataset is rather challenging due to large motions, complex lighting conditions, and thin object structures. To ensure a fair comparison, we downsampled images to 540 $\times$ 960 in our experiments and followed the training and validation camera split provided by~\cite{park2021hypernerf}. We conducted experiments on the four "vrig" scenes,  and we also provided results of the "interp" scenes in the supplementary material.


\subsection{Ablation Study}

\begin{figure}[h]
  \centering
  \includegraphics[width=0.4\textwidth]{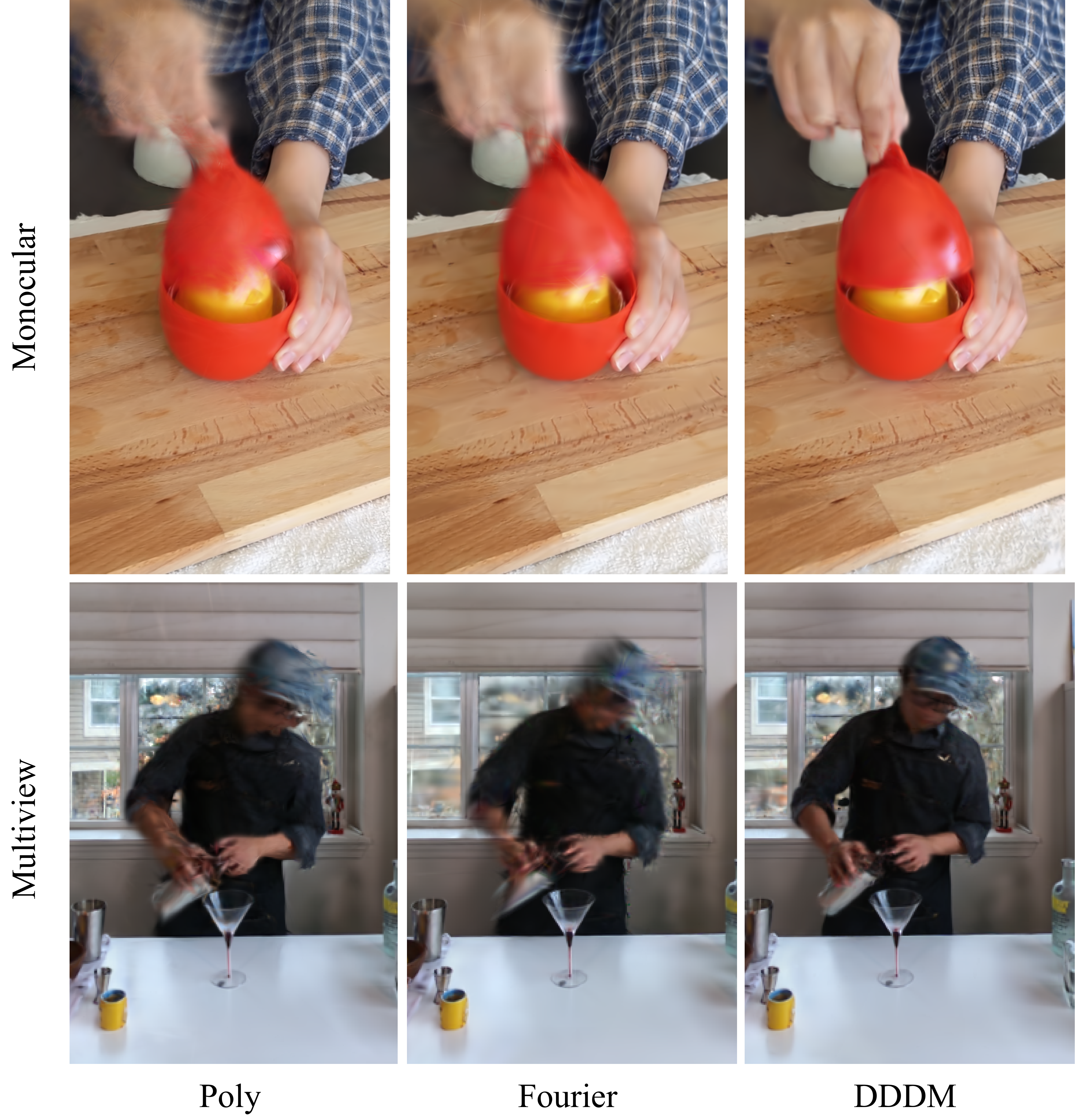}
  \caption{Ablation study on different deformation models. From left to right are deformation fitting with polynomial function only, Fourier series only, and our dual-domain deformation fitting. The proposed DDDM achieves the best rendering quality qualitatively.}
  \label{fig:fft-vs-poly}
\end{figure}

\paragraph{Deformation Models}
The deformation model is the core component of the proposed Gaussian-Flow. We conduct ablation studies to validate the effectiveness of the particular choice of DDDM. As DDDM consists of a Fourier series and a polynomial function, hence we first study the Fourier series and the polynomial functions separately, in which we only use the Fourier series or the polynomial function as the deformation model. As shown in Figure.~\ref{fig:fft-vs-poly}, the Fourier series contains more high-frequency components than the polynomial function, thus the Fourier series has sharper image details but results in more artifacts. The polynomial function is smoother than the Fourier series, leading to fewer artifacts but burry scene renderings. Finally, the hybrid DDDM function is able to generate sharper details with fewer artifacts.

\begin{figure}[h]
  \centering
  \includegraphics[width=0.45\textwidth]{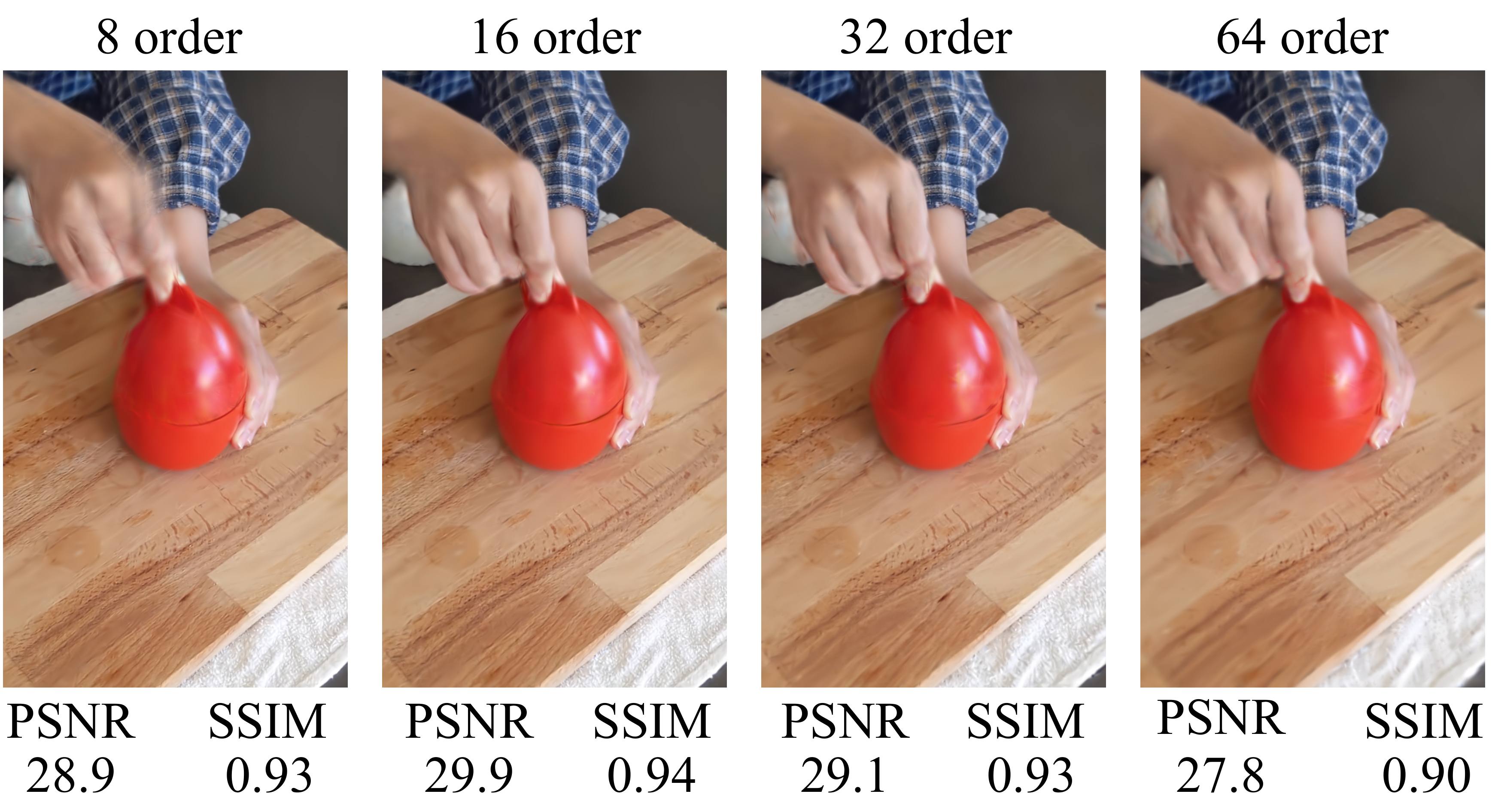}
  \caption{Ablation study on different orders of the DDDM. We find an order number of 16 leads to the best novel view rendering results in the HyperNeRF dataset.}
  \label{fig:dddm-order}
\end{figure}

Furthermore, we study the orders of the polynomial and Fourier series functions in our DDDM, which are related to the complexity of the scene and are crucial to the final performance. As shown in Figure.~\ref{fig:dddm-order}, the performance of our method increases with the number of Fourier series orders but starts to drop after the order over 32, which should be related to the over-parameterization of the deformation model.

\paragraph{Regularizations}
Next, we study the effectiveness of the two proposed regularizations in Gaussian-Flow optimization. As shown in Table.~\ref{tab:regularization}, adding separate KNN rigid regularization or the time smooth regularization can both improve the novel view rendering quality, and the full model that contains both regularizations can achieve the best performance.

\begin{table}[h]
  \centering
  \small
  \begin{tabular}{c|c|c|c}
  \toprule
  reg, & w/o KNN rigid & w/o Time smooth & full model  \\ 
  \midrule
  PSNR &  28.48& 29.12& 29.92\\
  SSIM &   0.92&  0.93&  0.94\\
  \bottomrule
  \end{tabular}
  \caption{Ablation study on the proposed KNN rigid and time smooth regularizations. The quantitative results demonstrate the effectiveness of both regularizations.}
  \label{tab:regularization}
  \end{table}




\subsection{Quantitative Comparisons}

\begin{table*}[ht]
    \caption{Per-scene quantitative comparisons on HyperNeRF~\cite{park2021hypernerf} dataset. Results are gathered from papers of the corresponding methods. Our method achieves the fastest training time, the highest rendering FPS, and the highest PSNR score for novel view synthesis, setting a new state-of-the-art for image-based dynamic scene rendering. }
    \vspace{-2mm}
    \begin{center}
    \scriptsize
    \begin{tabular}{l|c|c|cccccccc|cc}
    \toprule
    \multirow{2}{*}{\textbf{Method}} 
    & \multirow{2}{*}{\textbf{Train Time}$\downarrow$} 
    & \multirow{2}{*}{\textbf{Render FPS}$\uparrow$}
    & \multicolumn{2}{c}{\textbf{Broom}} & \multicolumn{2}{c}{\textbf{3D Printer}} 
    & \multicolumn{2}{c}{\textbf{Chicken}} & \multicolumn{2}{c}{\textbf{Peel Banana}} \vline 
    & \multicolumn{2}{c}{\textbf{Mean}} \\
    \cmidrule(lr){4-5} 
    \cmidrule(lr){6-7} \cmidrule(lr){8-9} \cmidrule(lr){10-11} \cmidrule(lr){12-13} 
    & &  
    & \textbf{PSNR}$\uparrow $ & \textbf{SSIM}  $\uparrow $ 
    & \textbf{PSNR}$\uparrow $  & \textbf{SSIM}  $\uparrow $ 
    & \textbf{PSNR}$\uparrow $  & \textbf{SSIM} $\uparrow $ 
    & \textbf{PSNR}$\uparrow $  & \textbf{SSIM} $\uparrow $ 
    & \textbf{PSNR}$\uparrow $  & \textbf{SSIM}$\uparrow $ \\
    \midrule
    NeRF~\cite{mildenhall2020nerf}
    & 16 hours
    &  0.013
    & 19.9&0.653 
    & 20.7&0.780 
    & 19.9&0.777
    & 20.0&0.769 
    & 20.1&0.745 \\
    Nerfies~\cite{nerfies}
    & 16 hours
    & 0.011
    & 19.2 &0.567 
    & 20.6&0.830 
    & 26.7& 0.943 
    & 22.4& 0.872 
    & 22.2& 0.803 \\
    HyperNeRF~\cite{park2021hypernerf}
    & 32 hours
    & 0.011
    & 19.3 &0.591 
    & 20.0& 0.821
    & 26.9& \cellcolor{sotacolor}0.948
    & 23.3& 0.896  
    & 22.4& 0.814 \\
    NeRFPlayer~\cite{song2022nerfplayer}
    & 6 hours
    & 0.208
    & 21.7 & 0.635
    & 22.9 & 0.810
    & 26.3 & 0.905
    & 24.0 & 0.863
    & 23.7 & 0.803 \\
    TiNeuVox~\cite{fang2022fast}
    & 30 min 
    & 0.5 
    & 21.5 & 0.686
    & 22.8 & 0.841
    & 28.3 & 0.947
    & 24.4 & 0.873
    & 24.3 & 0.837 \\
    \midrule
    Ours (30K)
    & 7 min
    & 125
    & 22.5 &0.690
    & 24.3 &0.857
    & 29.4 &0.934
    & 26.3 &0.906
    & 25.6 &0.847 \\
    Ours (60K)
    & \cellcolor{sotacolor}12 min
    & \cellcolor{sotacolor}125
    & \cellcolor{sotacolor}22.8 &\cellcolor{sotacolor}0.709
    & \cellcolor{sotacolor}25.0 &\cellcolor{sotacolor}0.877
    & \cellcolor{sotacolor}30.4 &0.945
    & \cellcolor{sotacolor}27.0 &\cellcolor{sotacolor}0.917
    & \cellcolor{sotacolor}26.3 &\cellcolor{sotacolor}0.862 \\
    \bottomrule
    \end{tabular}
    \end{center}
    \label{tab:hybernerf_bench}
    \end{table*}

We compare our method against previous SOTA NeRF-based methods, including NeRF~\cite{mildenhall2020nerf}, Nerfies~\cite{nerfies}, HyperNeRF~\cite{park2021hypernerf}, NeRFPlayer~\cite{song2022nerfplayer}, and TiNeuVox~\cite{fang2022fast}. We also provide comparisons with other 3DGS-based approaches concurrently proposed with our Gaussian-Flow. The training time, rendering FPS, and novel view synthesis PSNR of different methods can be found in Table.~\ref{tab:hybernerf_bench}. Previous NeRF-based methods require at least 30 minutes to train the scene, and fail to achieve real-time rendering of the dynamic scene. Our method only requires \textbf{7 minutes} of training time and can achieve real-time rendering speed, which is much faster than previous methods. Moreover, our method can achieve better performance than previous SOTA methods in terms of PSNR.

\begin{table}[ht]
    \caption{Quantitative comparison on Plenoptic Video dataset~\cite{Li_2022_CVPR}. Our training speed is $5\times$ faster of magnitude faster than previous leading approaches, . Also, we achieved the highest PSNR score among all methods.}
    \vspace{-2mm}
    \begin{center}
    \small 
    \begin{tabularx}{\linewidth}{ X | c | r r }
    \toprule
    \textbf{Method} & \textbf{Train Time} $\downarrow$ & \textbf{PSNR} $\uparrow$ & \textbf{SSIM} $\uparrow$\\
    \midrule
    LLFF~\cite{llff} & - & 23.2 & - \\
    DyNeRF~\cite{Li_2022_CVPR} & 1344 hours & 29.6 & 0.96 \\
    NeRFPlayer~\cite{song2022nerfplayer} & 5.5 hours & 30.7 & - \\
    K-Planes~\cite{fridovich2023k} & 1.8 hours & 31.6 & 0.96 \\
    4D-GS~\cite{wu20234d}   & 2 hours & 31.0 & 0.94 \\
    \midrule
    Ours (30K) &  22.5 min & 30.5 & 0.97 \\
    Ours (60K) & \cellcolor{sotacolor} 41.8 min & \cellcolor{sotacolor}32.0 & \cellcolor{sotacolor}0.97 \\
    \bottomrule
    \end{tabularx}
    \end{center}
    \label{tab:3dv_bench}
    \end{table}

We evaluated various methods on the Plenoptic Video dataset, as summarized in Table.~\ref{tab:3dv_bench}. The comparison focuses on training time efficiency and image quality, assessed through PSNR and SSIM. Our approach demonstrated a significant advancement in training efficiency, requiring only 22.5 minutes, a drastic reduction compared to the hours needed by methods like DyNeRF~\cite{Li_2022_CVPR} and K-Planes~\cite{fridovich2023k}. This efficiency is paramount for practical applications, where reduced training time can be a critical factor. In terms of image quality, our method achieved a PSNR of 30.5 with 30K steps, which, while not the highest, is competitive with the leading methods. However, our method scored 0.97 in SSIM, higher than K-Planes' leading score of 0.96. This indicates a potential trade-off between training efficiency and the ability to preserve structural details in images. 

In addition, we extend our method to 60K steps on both datasets, which can further improve the performance on the HyperNeRF dataset~\cite{park2021hypernerf}, and achieve the highest performance on the Plenoptic Video dataset~\cite{Li_2022_CVPR}. However, the training time is also increased by approximately 2$\times$ for HyperNeRF dataset~\cite{park2021hypernerf} and about 1.5$\times$ for Plenoptic Video dataset~\cite{Li_2022_CVPR}, which is still much faster than previous methods.

\subsection{Qualitative Comparisons}

\begin{figure*}[htb]
    \centering
    \begin{subfigure}[b]{0.145\linewidth}
        \includegraphics[width=\linewidth]{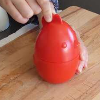}
    \end{subfigure}
    \begin{subfigure}[b]{0.145\linewidth}
        \includegraphics[width=\linewidth]{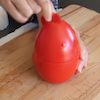}
    \end{subfigure}
    \begin{subfigure}[b]{0.145\linewidth}
        \includegraphics[width=\linewidth]{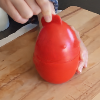}
    \end{subfigure}
    \begin{subfigure}[b]{0.145\linewidth}
        \includegraphics[width=\linewidth]{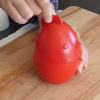}
    \end{subfigure}
    \begin{subfigure}[b]{0.145\linewidth}
        \includegraphics[width=\linewidth]{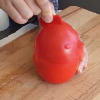}
    \end{subfigure}
    \begin{subfigure}[b]{0.145\linewidth}
        \includegraphics[width=\linewidth]{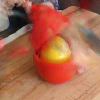}
    \end{subfigure}
    \begin{subfigure}[b]{0.074\linewidth}
        \includegraphics[width=\linewidth]{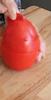}
    \end{subfigure}

    \begin{subfigure}[b]{0.145\linewidth}
        \includegraphics[width=\linewidth]{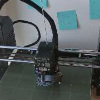}
        \caption{GT}
        \caption{ }
    \end{subfigure}
    \begin{subfigure}[b]{0.145\linewidth}
        \includegraphics[width=\linewidth]{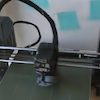}
        \caption{Ours (30K)}
        \caption{(12 min)}
    \end{subfigure}
    \begin{subfigure}[b]{0.145\linewidth}
        \includegraphics[width=\linewidth]{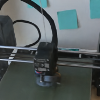}
        \caption{TiNeuVox}
        \caption{(30 hours)}
    \end{subfigure}
    \begin{subfigure}[b]{0.145\linewidth}
        \includegraphics[width=\linewidth]{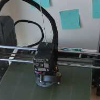}
        \caption{HyperNeRF}
        \caption{(32 hours)}
    \end{subfigure}
    \begin{subfigure}[b]{0.145\linewidth}
        \includegraphics[width=\linewidth]{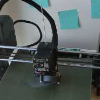}
        \caption{Nerfies}
        \caption{(16 hours)}
    \end{subfigure}
    \begin{subfigure}[b]{0.145\linewidth}
        \includegraphics[width=\linewidth]{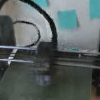}
        \caption{NeRF}
        \caption{(16 hours)}
    \end{subfigure}
    \begin{subfigure}[b]{0.074\linewidth}
        \includegraphics[width=\linewidth]{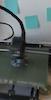}
        \caption{NP}
        \caption{(6 hours)}
    \end{subfigure}

\caption{Qualitative comparisons of our method and TiNeuVox~\cite{fang2022fast}, HyperNeRF~\cite{park2021hypernerf}, Nerfies~\cite{nerfies} and NeRFPlayer (NP)~\cite{fridovich2023k} on the HyperNeRF~\cite{park2021hypernerf} dataset. The training time of each method is shown in the brackets.}
\label{fig:hybernerf_bench}
\end{figure*}

In this section, we show qualitative comparisons of our method and previous SOTA methods on the HyperNeRF~\cite{park2021hypernerf} dataset and the Plenoptic Video dataset~\cite{Li_2022_CVPR}. Figure.~\ref{fig:hybernerf_bench} shows the qualitative comparisons of our method and TiNeuVox~\cite{fang2022fast}, HyperNeRF~\cite{park2021hypernerf}, Nerfies~\cite{nerfies} and NeRFPlayer~\cite{fridovich2023k} on HyperNeRF dataset. Notice that our method can produce comparably clear and sharp images than previous SOTA methods, which highlights our method's superior performance in monocular conditions. Despite its overall efficacy, our method does encounter limitations in extremely thin structure, as shown in the 3D Printer scene, the thread of the 3D Printer is not clear, while other methods can produce a clear thread.

\begin{figure*}[ht]
    \centering
    \begin{subfigure}[b]{0.31\linewidth}
        \includegraphics[width=\linewidth]{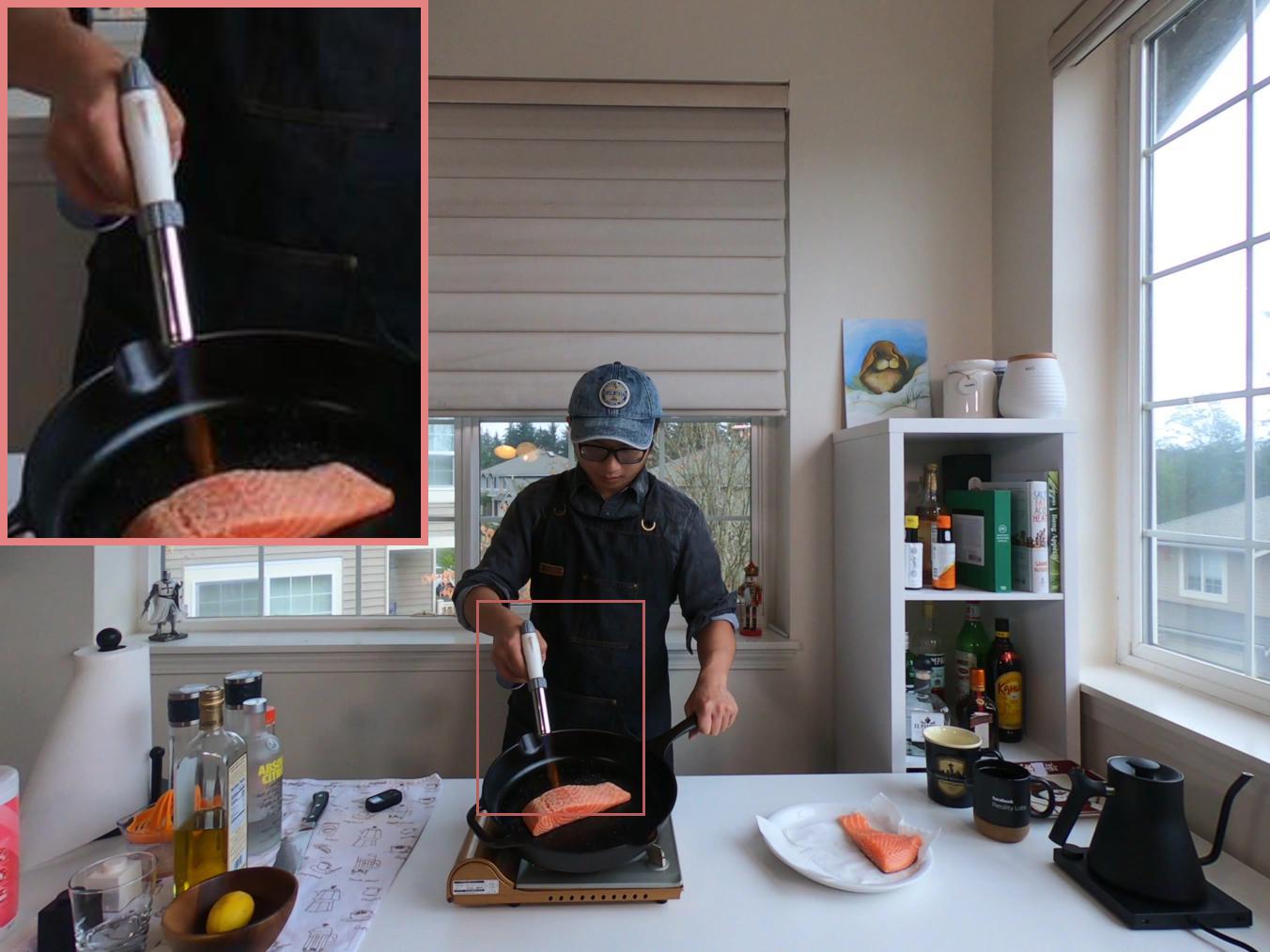}
        \caption{GT}
    \end{subfigure}
    \begin{subfigure}[b]{0.31\linewidth}
        \includegraphics[width=\linewidth]{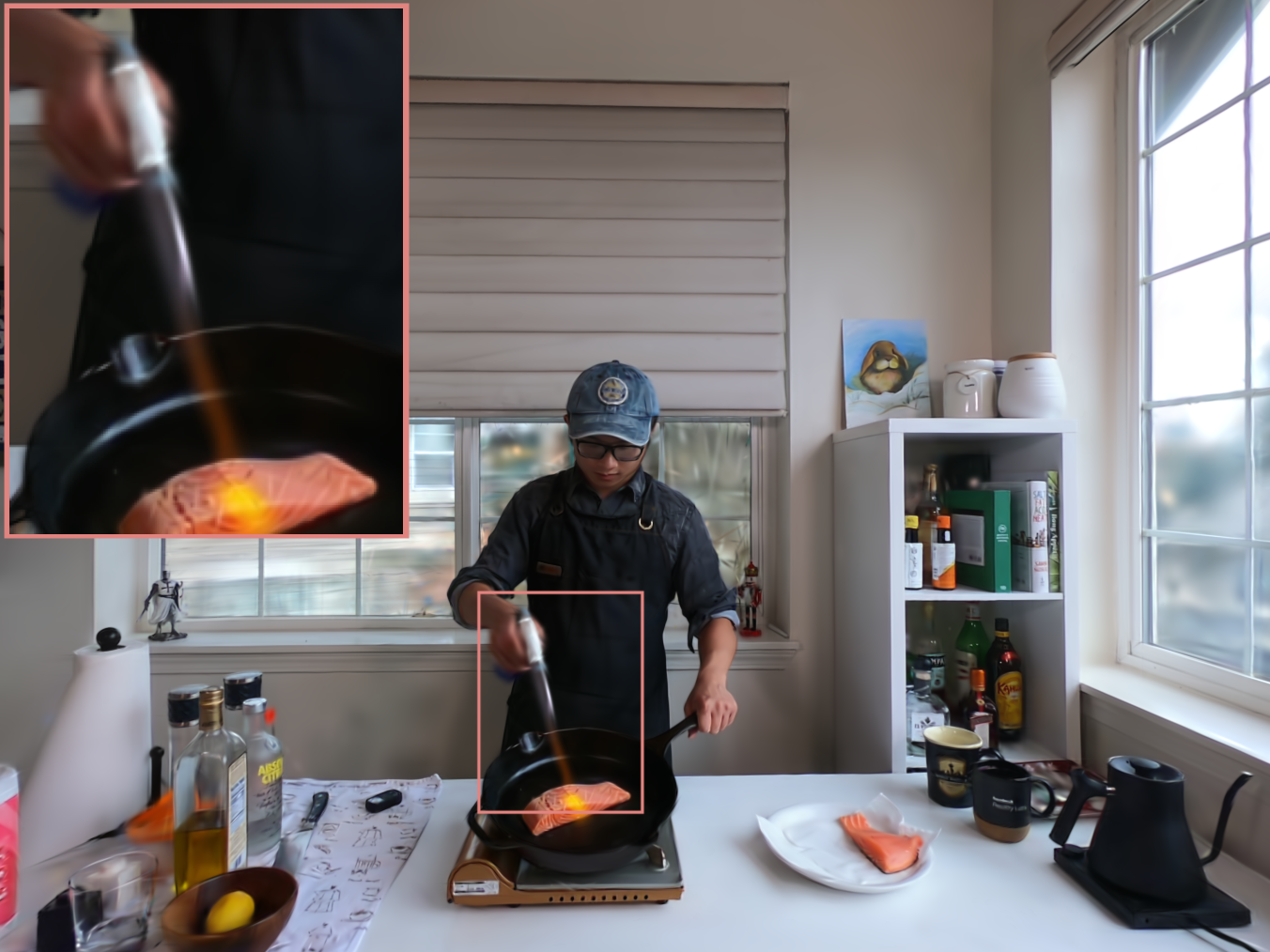}
        \caption{Ours (30K)}
    \end{subfigure}
    \begin{subfigure}[b]{0.31\linewidth}
        \includegraphics[width=\linewidth]{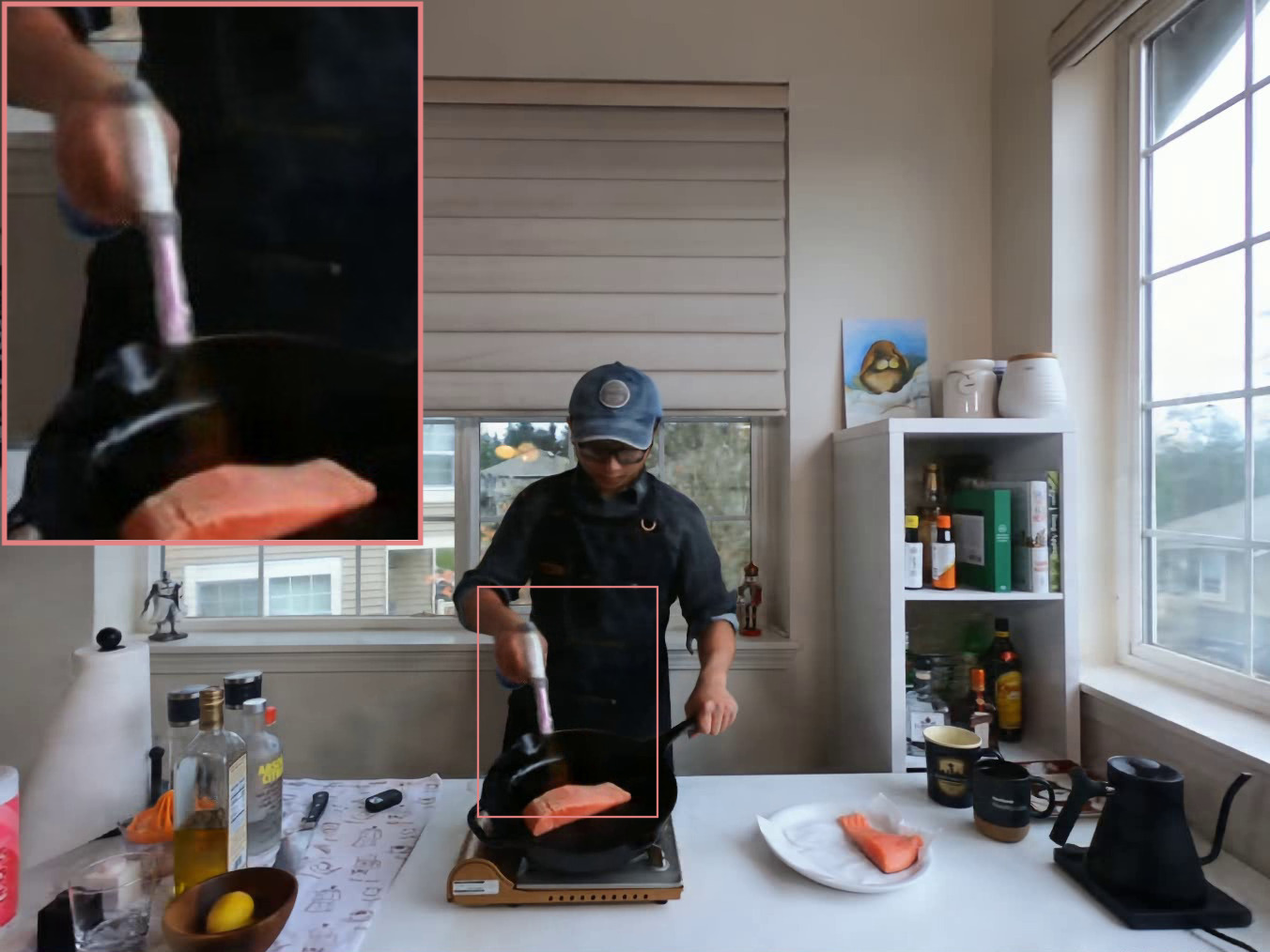}
        \caption{K-Planes}
    \end{subfigure}
    \begin{subfigure}[b]{0.31\linewidth}
        \includegraphics[width=\linewidth]{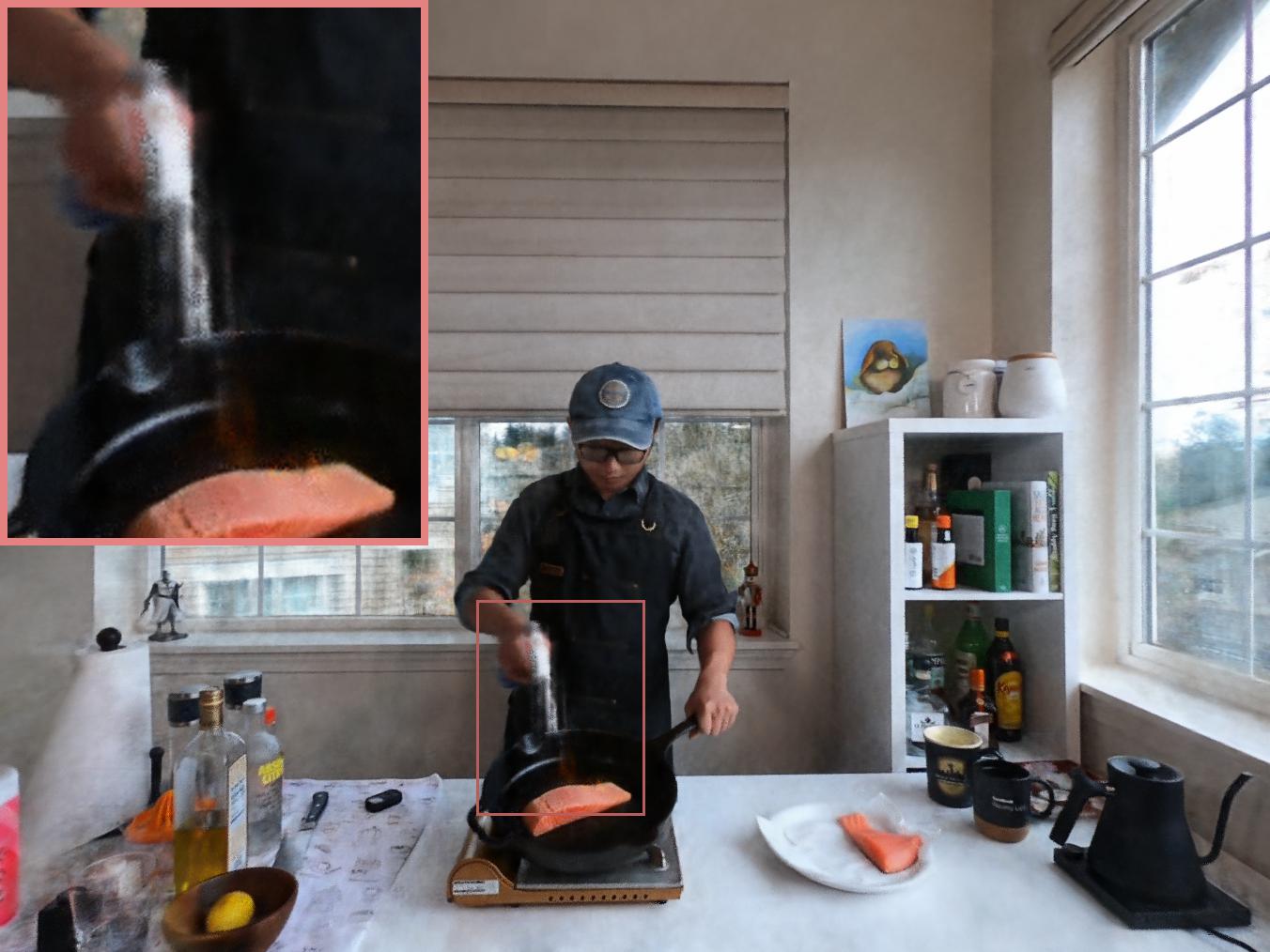}
        \caption{NeRFPlayer}
    \end{subfigure}
    \begin{subfigure}[b]{0.31\linewidth}
        \includegraphics[width=\linewidth]{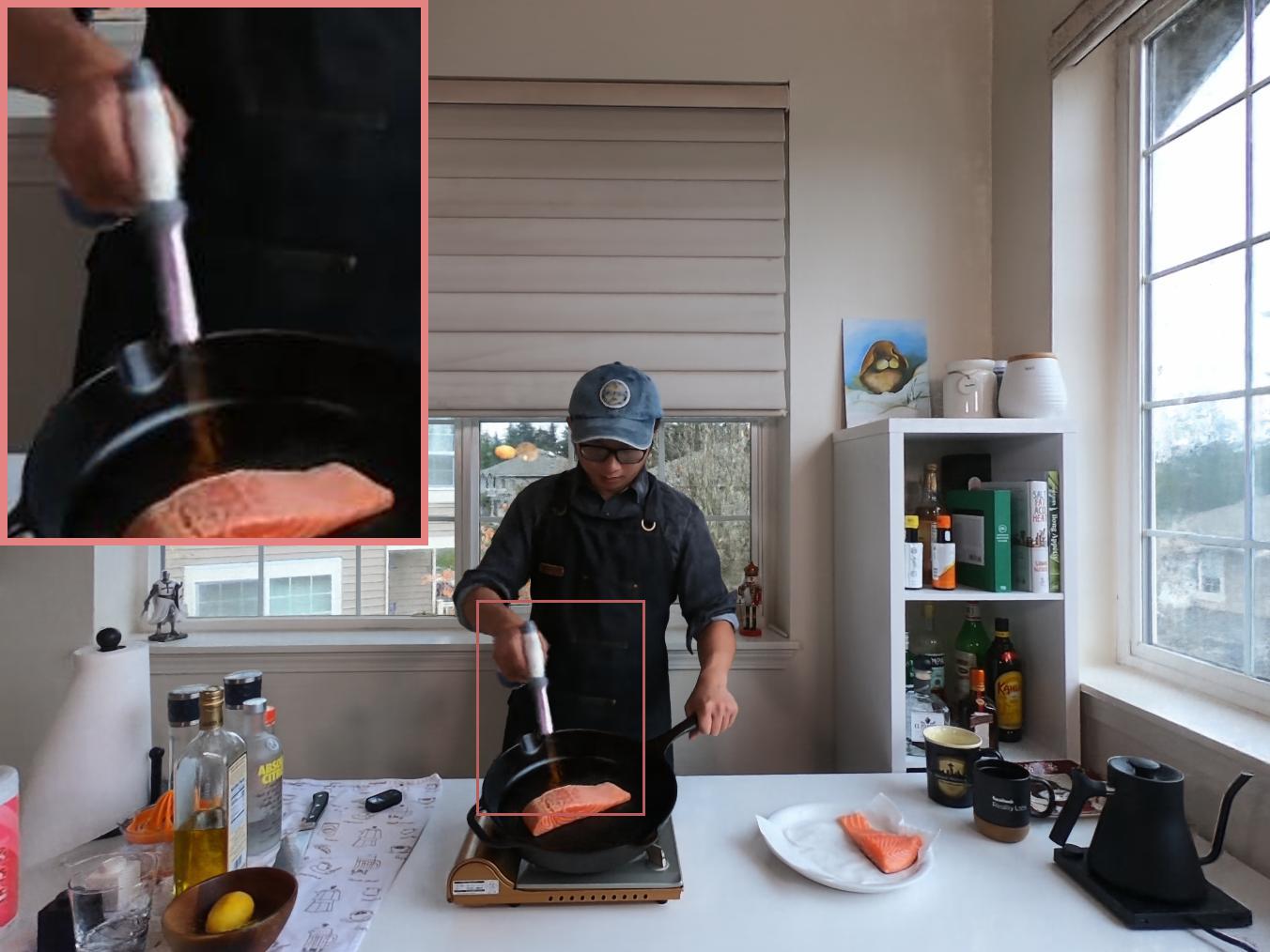}
        \caption{DyNeRF}
    \end{subfigure}
    \begin{subfigure}[b]{0.31\linewidth}
        \includegraphics[width=\linewidth]{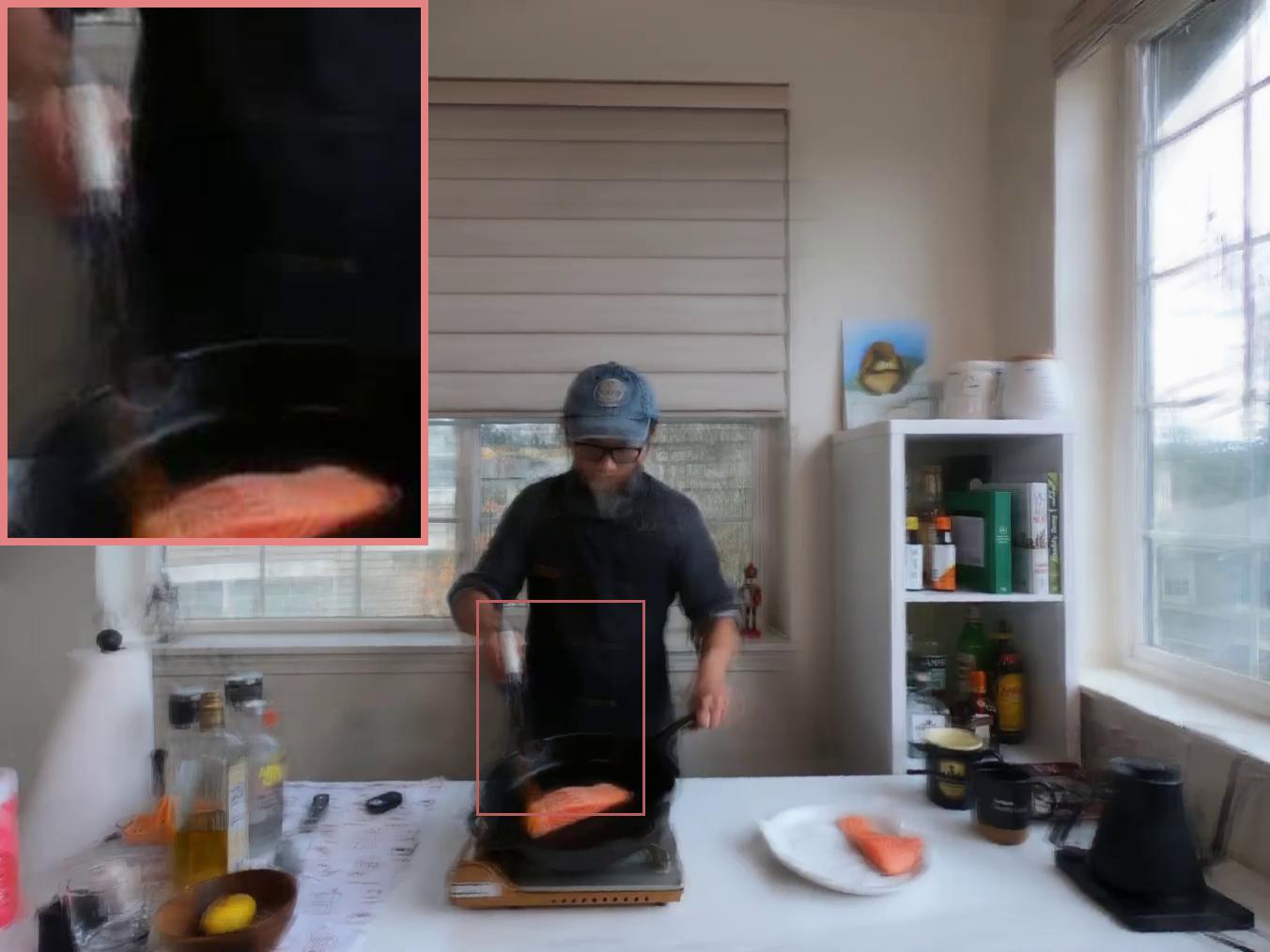}
        \caption{LLFF}
    \end{subfigure}
    \caption{Qualitative comparisons of our method and K-Planes~\cite{fridovich2023k}, NeRFPlayer~\cite{fridovich2023k}, DyNeRF~\cite{Li_2022_CVPR}, and LLFF on Plenoptic Video dataset.}
    \label{fig:3dv_bench}
    \end{figure*}

We also compare our method with previous SOTA methods on the Plenoptic Video dataset~\cite{Li_2022_CVPR}, as shown in Figure.~\ref{fig:3dv_bench}. Compared with previous SOTA methods, our method can produce more accurate color and correct structure. Moreover, our method successfully reconstructs the flame in the scene, while NeRFPlayer~\cite{fridovich2023k} fails to reconstruct the flame. These results suggest that our method can achieve comparable image quality with previous SOTA methods, which demonstrates the effectiveness of our method in the multiview conditions. 

\captionsetup[subfigure]{labelformat=empty}
\captionsetup[subfigure]{labelformat=default}

\captionsetup[subfigure]{labelformat=empty}
\captionsetup[subfigure]{labelformat=default}

\section{Conclusion}

In this paper, we introduced Gaussian-Flow, a novel framework for dynamic 3D scene reconstruction using a point-based differentiable rendering approach. The core of our innovation lies in the DDDM, which efficiently models deformations of each 3D Gaussian point in both the time and frequency domains. This approach has enabled us to set a new state-of-the-art for 4D scene reconstruction in terms of training speed, rendering frames per second, and novel view synthesis quality. Our extensive experiments and ablation studies have demonstrated the efficacy of proposed Gaussian-Flow across various datasets. We achieved significant improvements over existing methods, particularly in training speed and rendering performance. The ability to efficiently handle dynamic scenes without the computational overhead of neural networks marks a substantial leap forward in this domain. 

\section{Limitations}

While our method excels in rendering speed and training efficiency, there is room for improvement in maintaining high-fidelity thin structures in the final rendering. Future work could focus on enhancing the balance between speed and image detail preservation, potentially through more refined deformation models or advanced regularization techniques.

{
    \small
    \bibliographystyle{ieeenat_fullname}
    \bibliography{main}
}

\clearpage
\setcounter{page}{1}
\maketitlesupplementary

\renewcommand{\thesection}{\Alph{section}}
\setcounter{section}{0}

\section{Implementation Details}

In this paper, we focus exclusively on modeling three key attributes of the 3D Gaussian Splatting (3DGS) using our novel DDDM model. These attributes are: 1) the position of the Gaussian, 2) the rotation represented by a quaternion, and 3) the first three coefficients of the Spherical Harmonics (SHs). The learning approach employed for each of these attributes within the DDDM framework mirrors that of the corresponding 3DGS attribute, ensuring consistency in our modeling strategy.

We first train each scene with no deformation (as a 3DGS) for 2000 iterations, and then train the scene with deformation (with DDDM) for the remaining training phase. We stop the process of adding (through splitting and cloning as delineated in 3DGS) and pruning Gaussian points as 15K iterations. Then, We start using the KNN rigid loss at 5000 iterations, since the number of Gaussian points is fixed, which is more computationally efficient, because we can only compute the KNN index once instead of calculating the KNN for each iteration.

\section{More Results}

This section presents additional visual results. These include a broader range of viewpoints and scenes, highlighting the capability of our method in rendering novel view variants across both spatial and temporal dimensions. We also showcase the proficiency of our approach in reconstructing depth maps. 

\begin{figure}[h]
    \centering
    \begin{tabular}{@{}c@{\hspace{0.25mm}}c@{\hspace{0.25mm}}c@{\hspace{0.25mm}}c@{}}
        \includegraphics[width=0.14\textwidth]{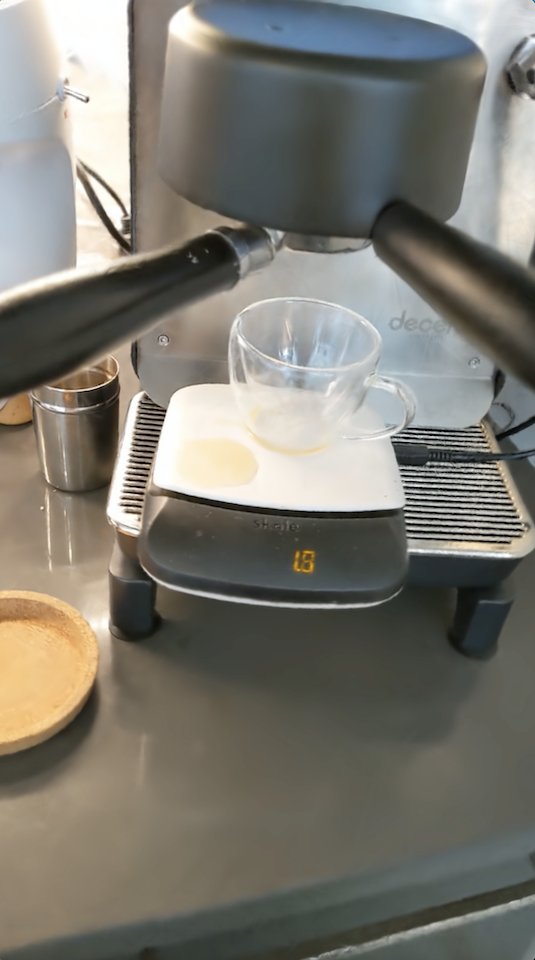} &
        \includegraphics[width=0.14\textwidth]{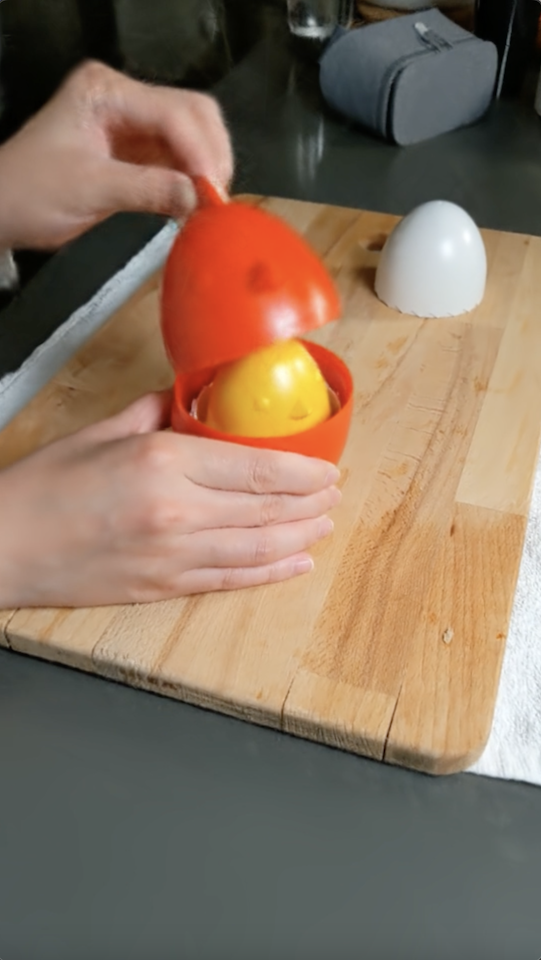}  &
        \includegraphics[width=0.14\textwidth]{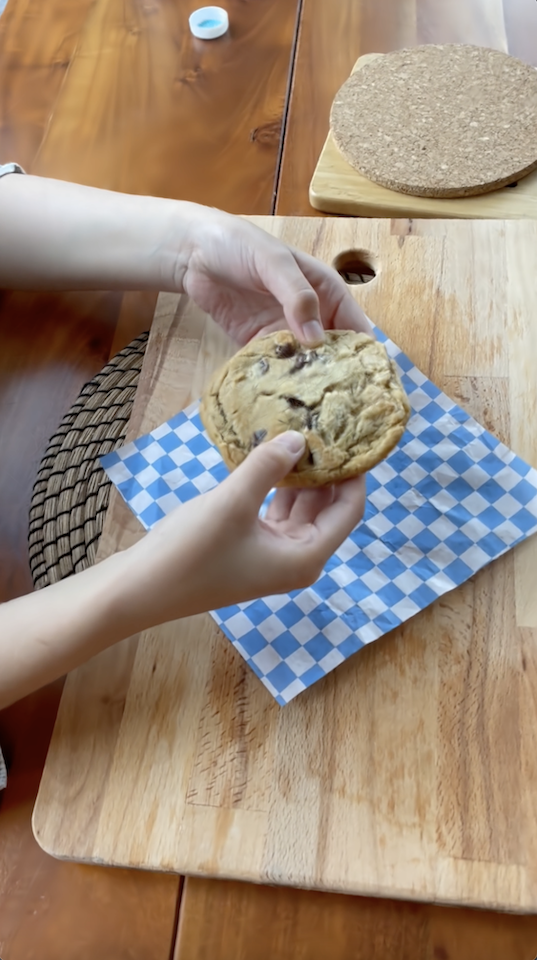} \\
        \includegraphics[width=0.14\textwidth]{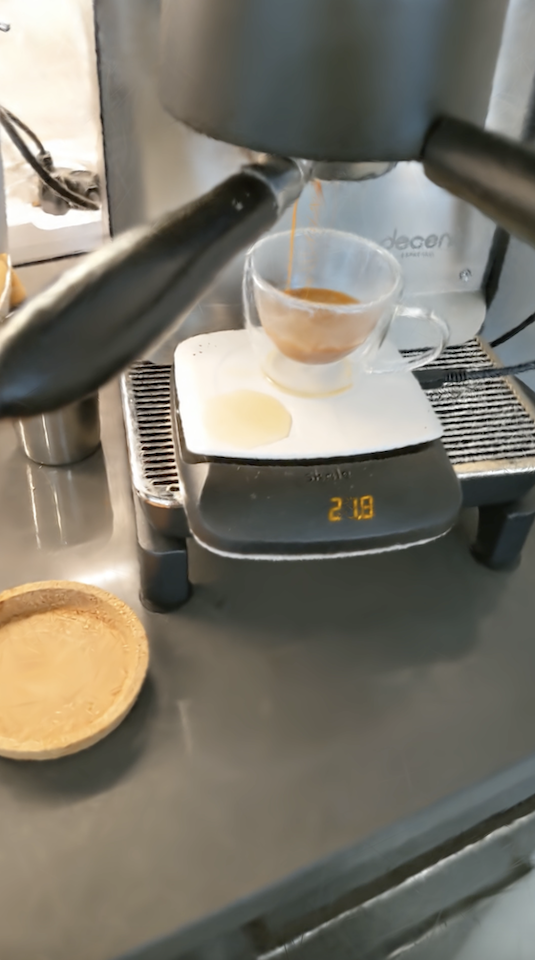} &
        \includegraphics[width=0.14\textwidth]{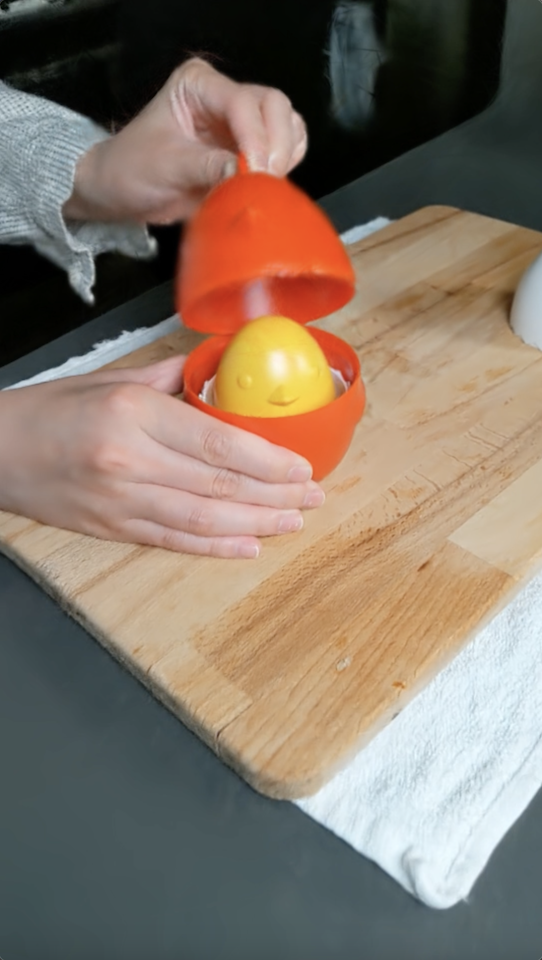}  &
        \includegraphics[width=0.14\textwidth]{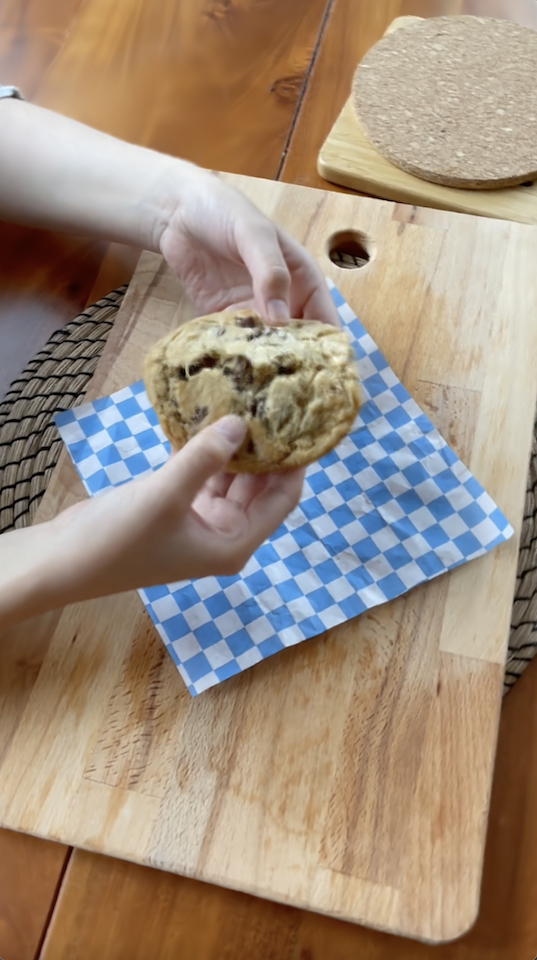} \\
        \includegraphics[width=0.14\textwidth]{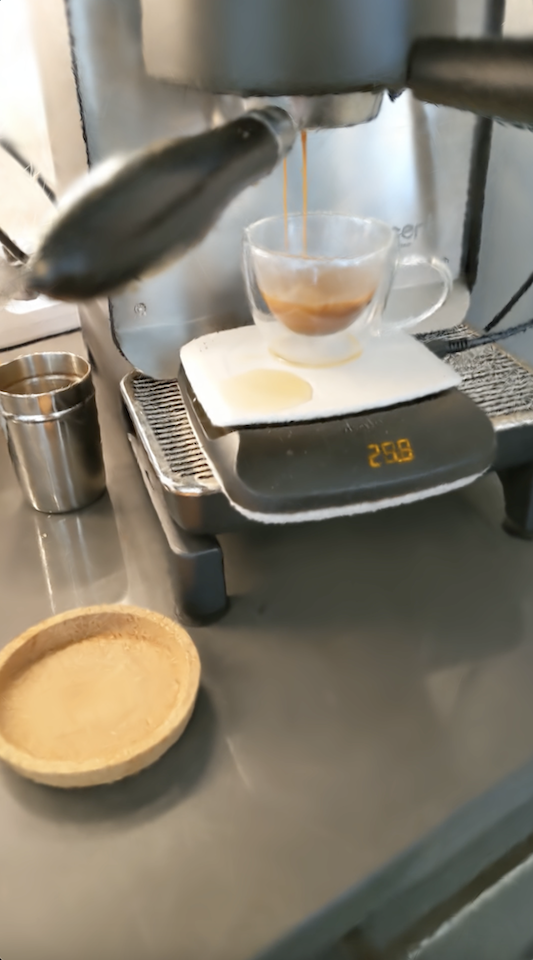} &
        \includegraphics[width=0.14\textwidth]{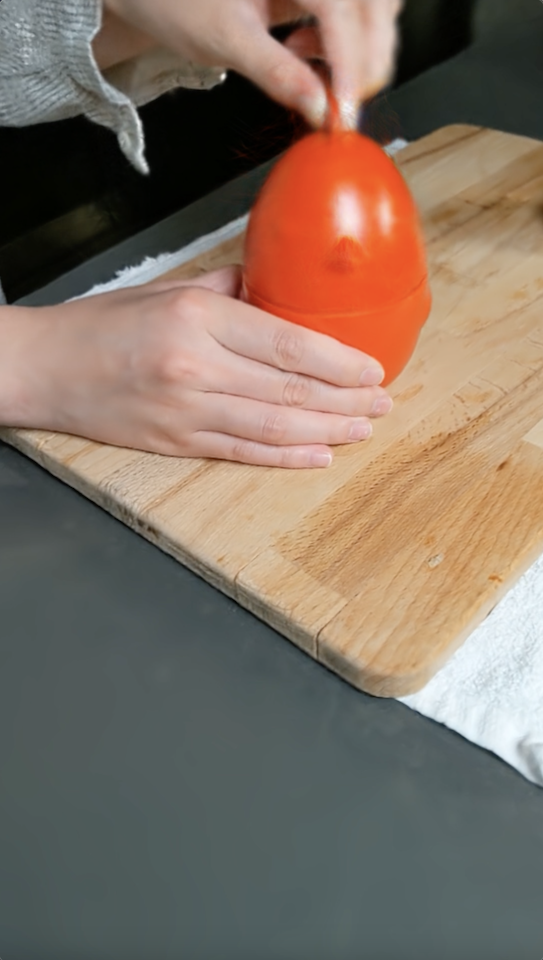}  &
        \includegraphics[width=0.14\textwidth]{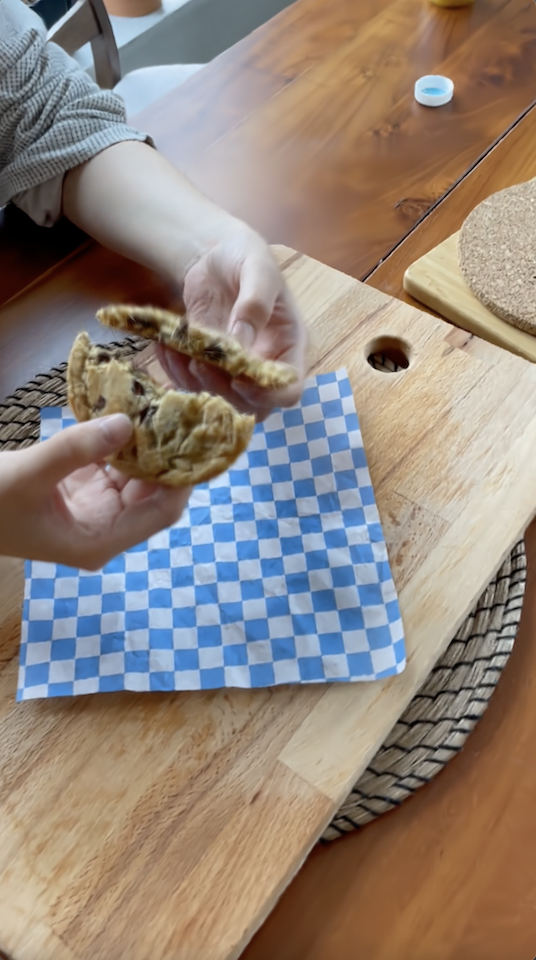} \\
        \includegraphics[width=0.14\textwidth]{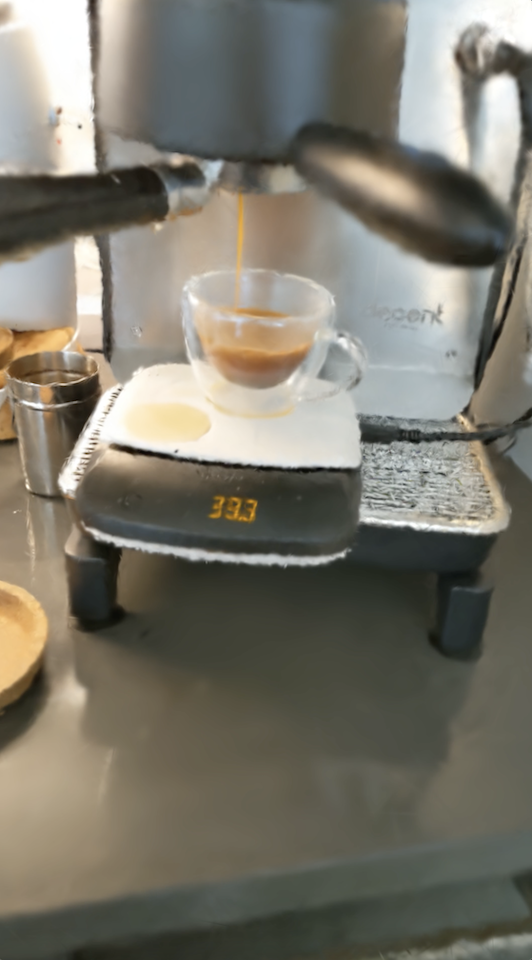} &
        \includegraphics[width=0.14\textwidth]{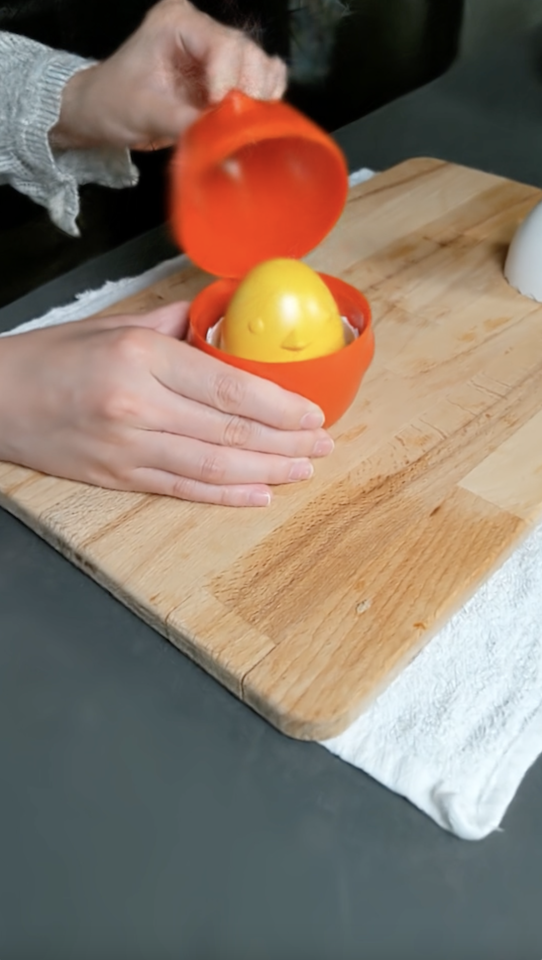}  &
        \includegraphics[width=0.14\textwidth]{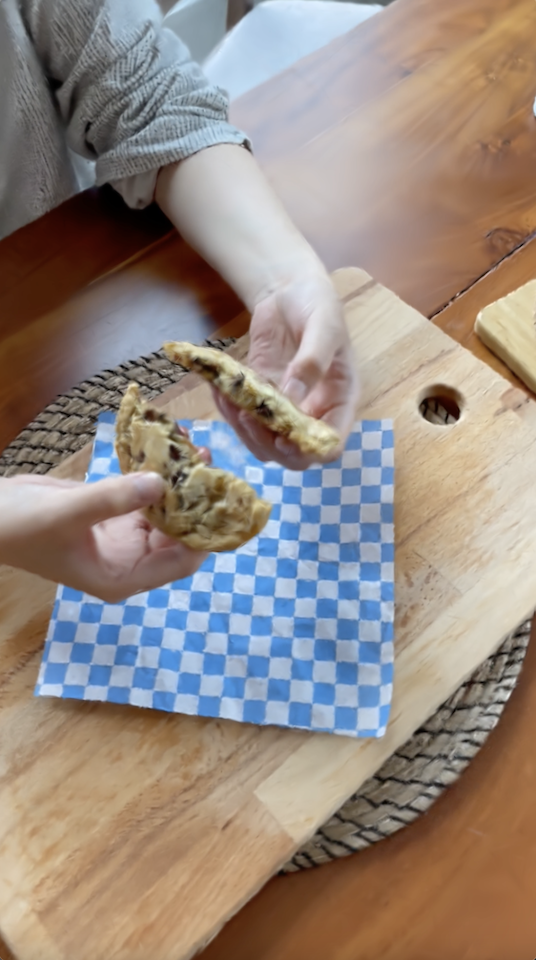} \\
    \end{tabular}
    \caption{
        \textbf{View Synthesis Results on HyperNeRF Dataset.}
        }
    \label{sup:inter_hyper}
    \vspace{7mm}
\end{figure}

\begin{figure*}[!htp]
    \centering
    \begin{tabular}{@{}c@{\hspace{0.25mm}}c@{\hspace{0.25mm}}c@{\hspace{0.25mm}}c@{\hspace{0.25mm}}c@{}}
        \includegraphics[width=0.25\textwidth]{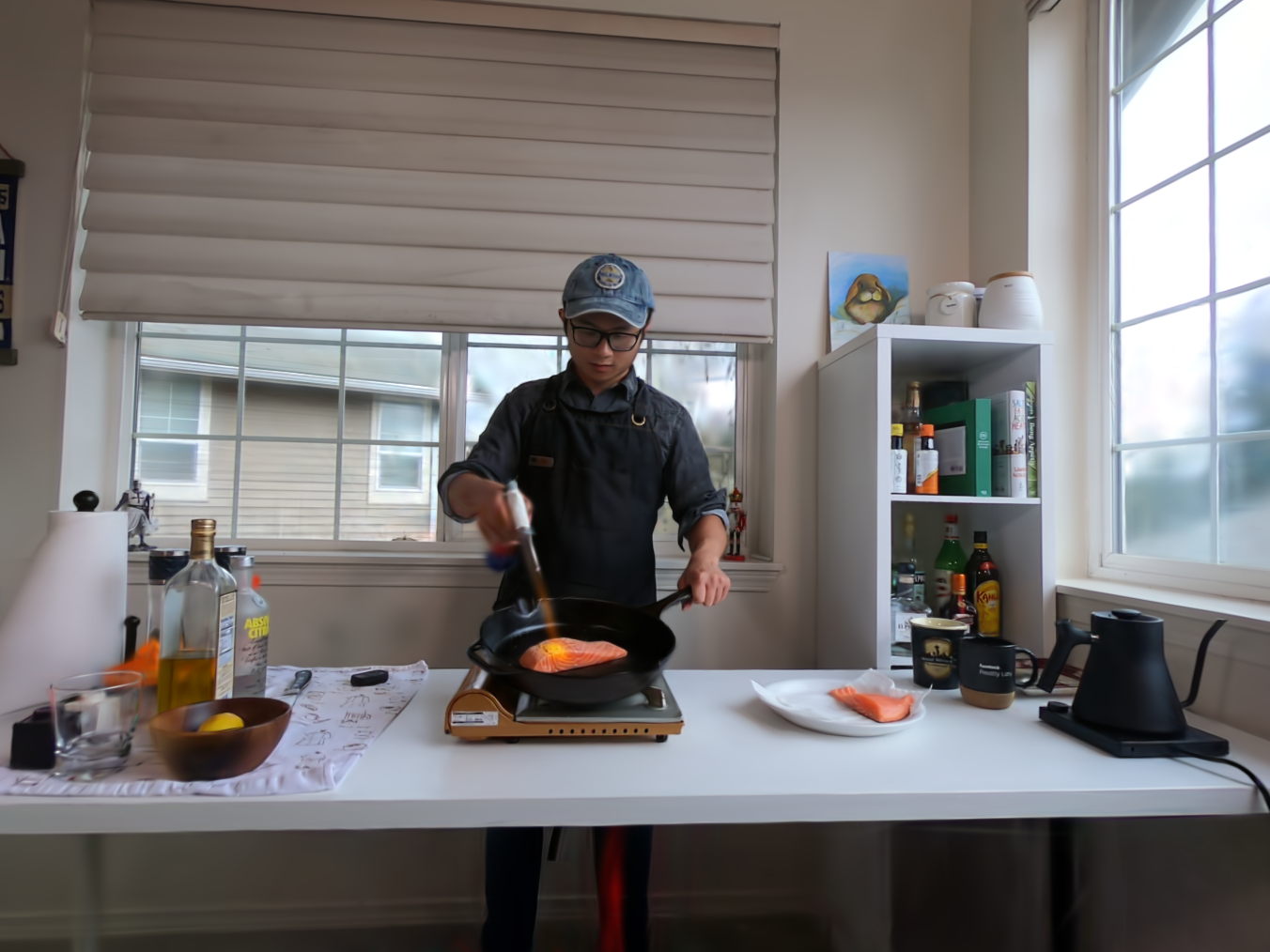} &
        \includegraphics[width=0.25\textwidth]{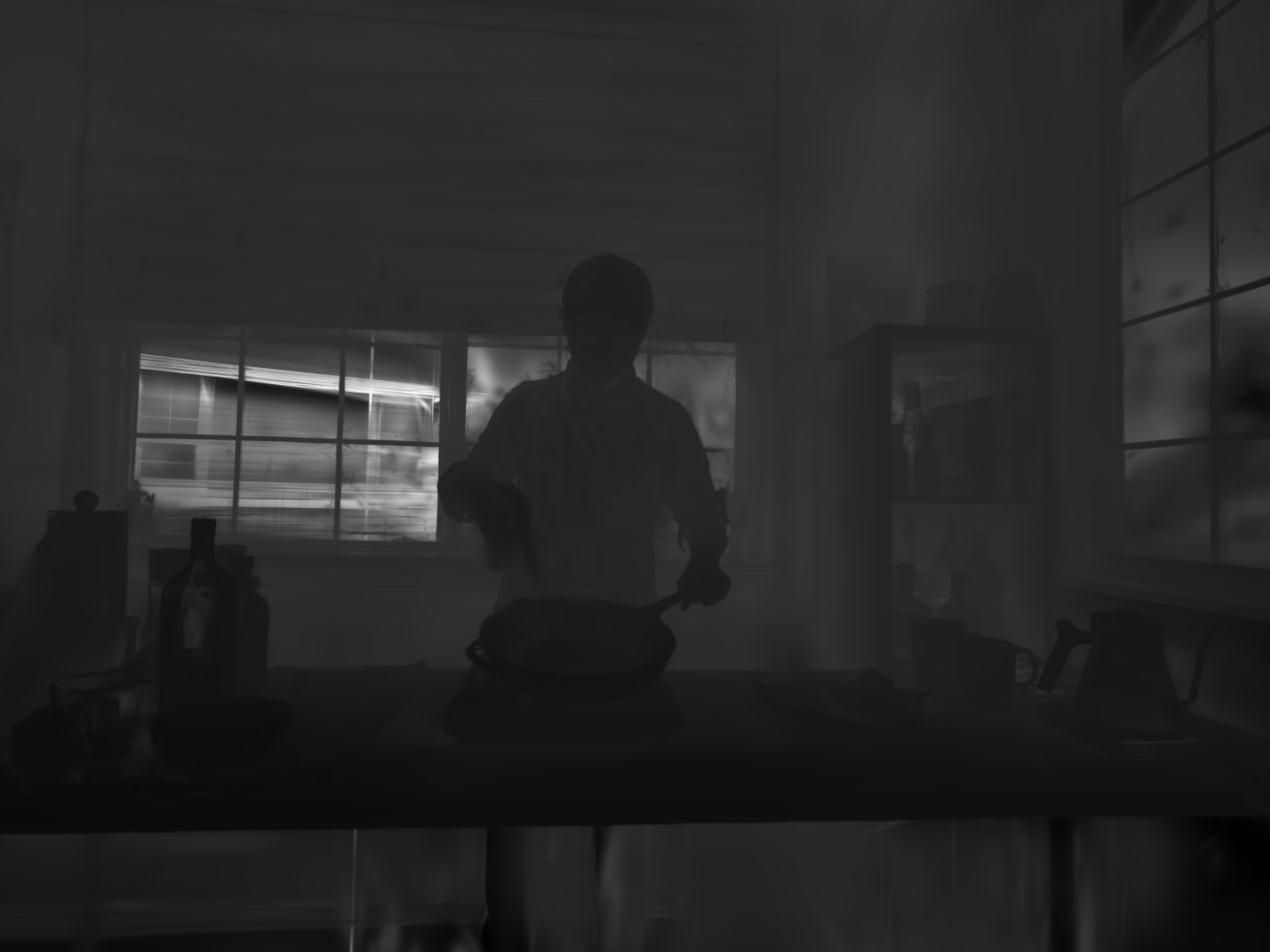} &
        \includegraphics[width=0.25\textwidth]{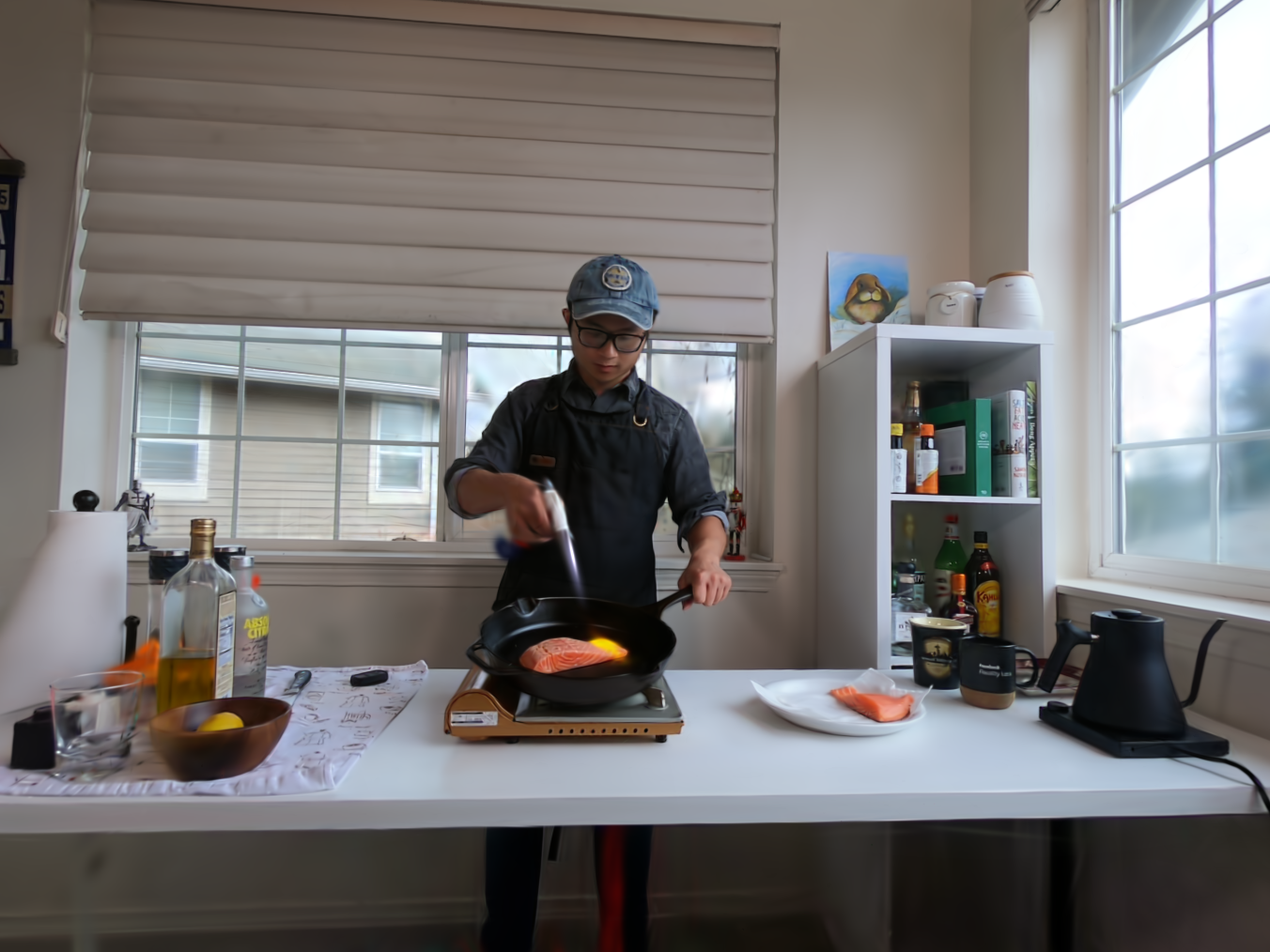} &
        \includegraphics[width=0.25\textwidth]{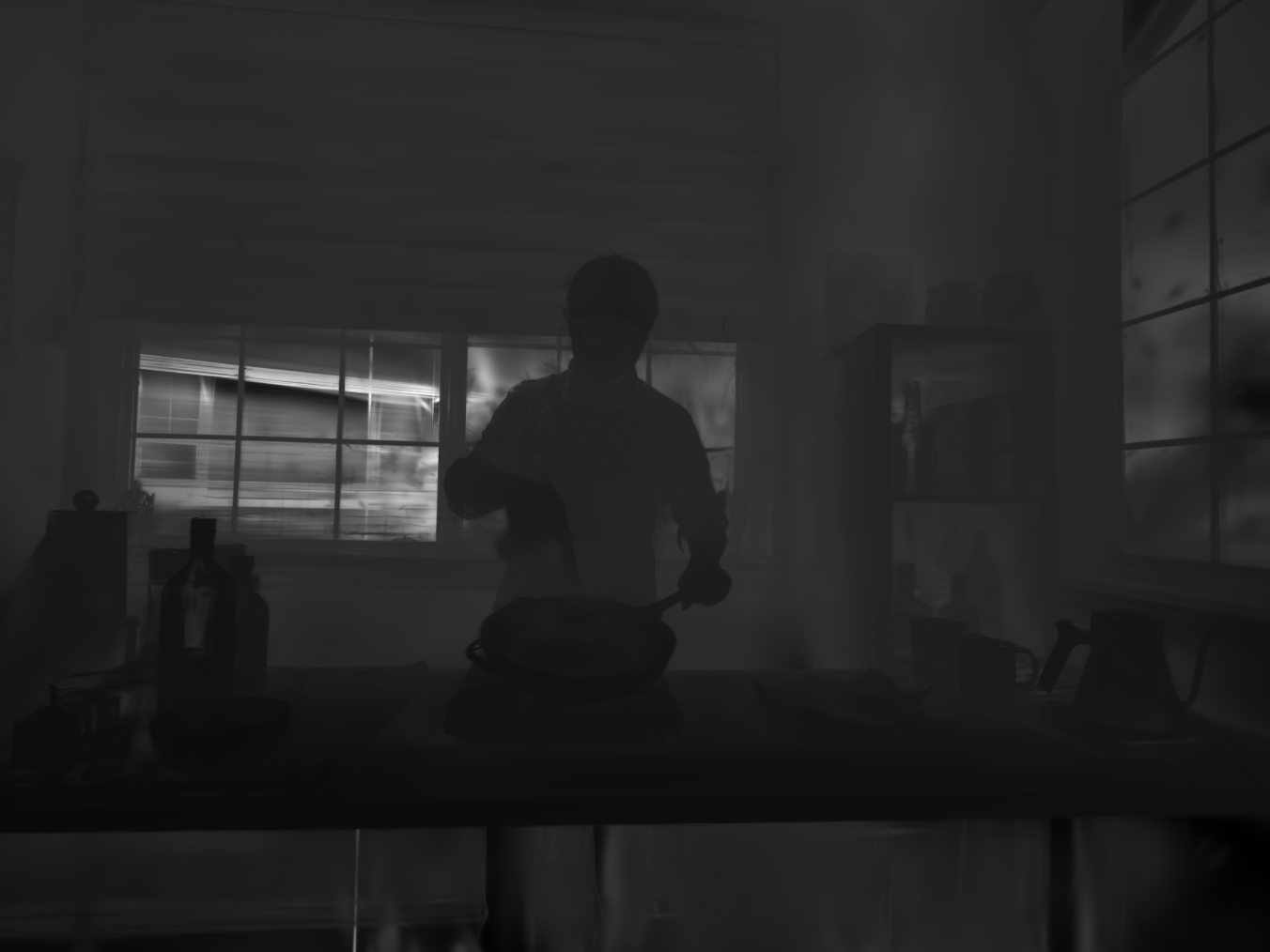} \\
        \includegraphics[width=0.25\textwidth]{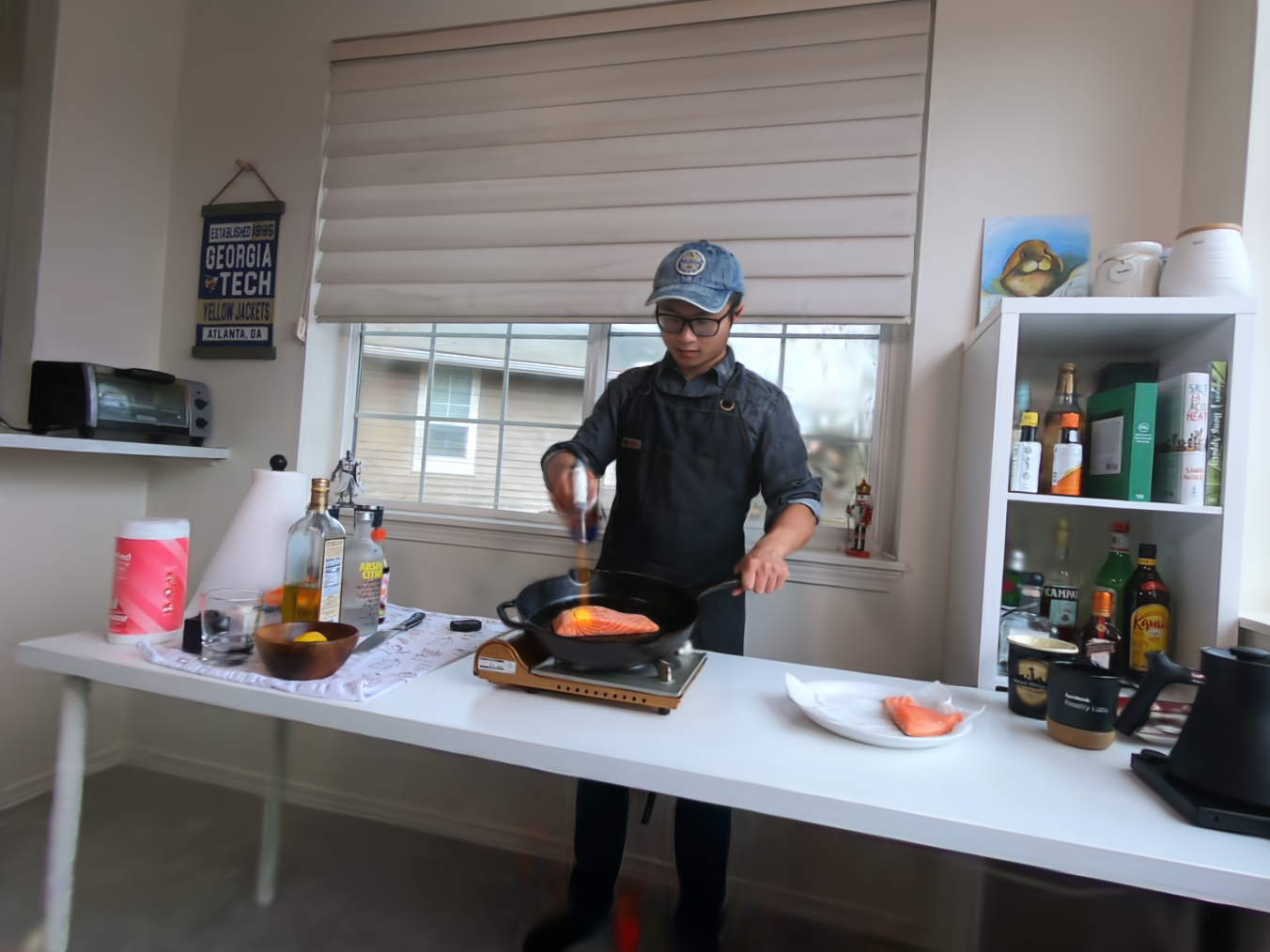} &
        \includegraphics[width=0.25\textwidth]{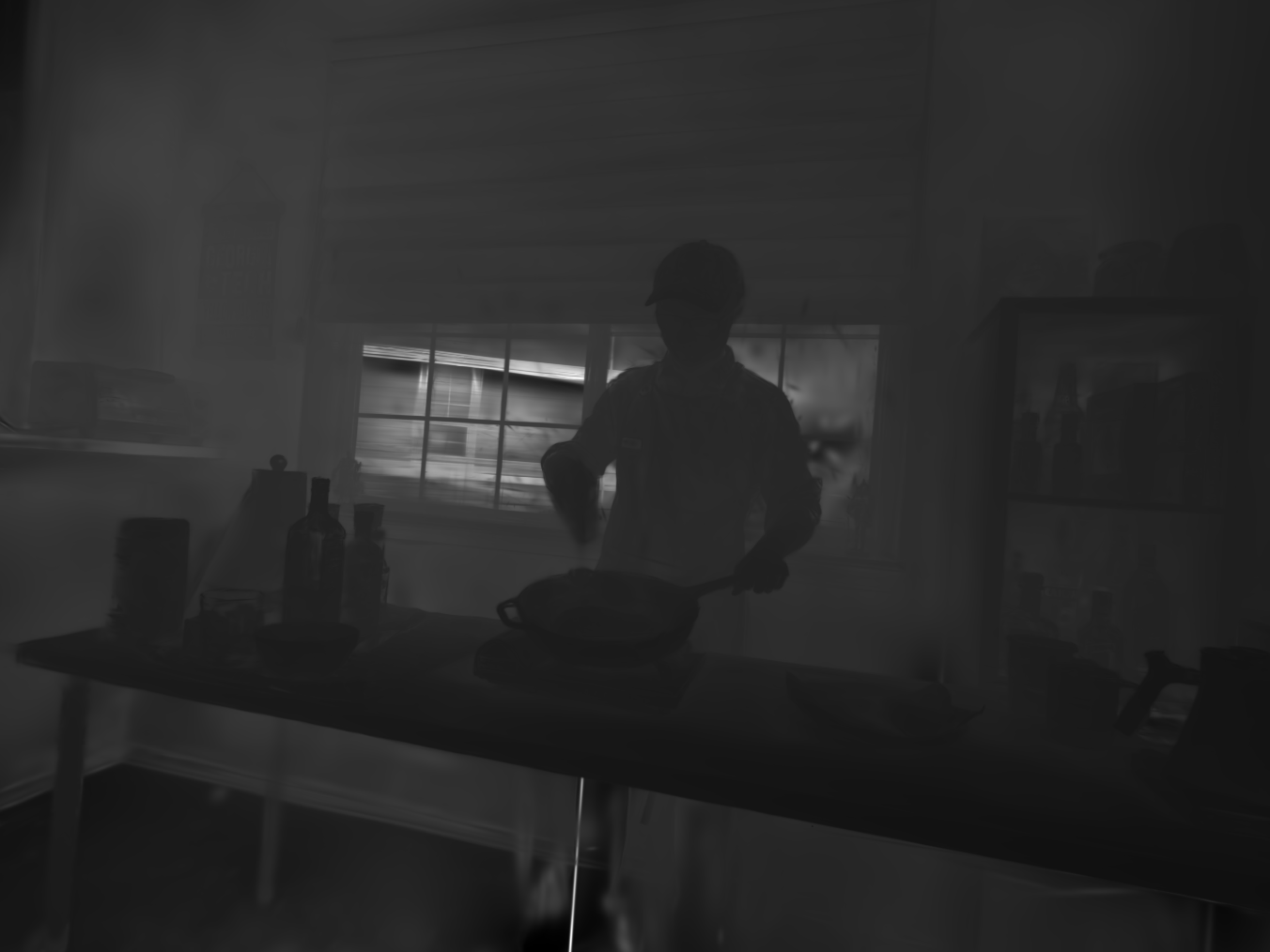} &
        \includegraphics[width=0.25\textwidth]{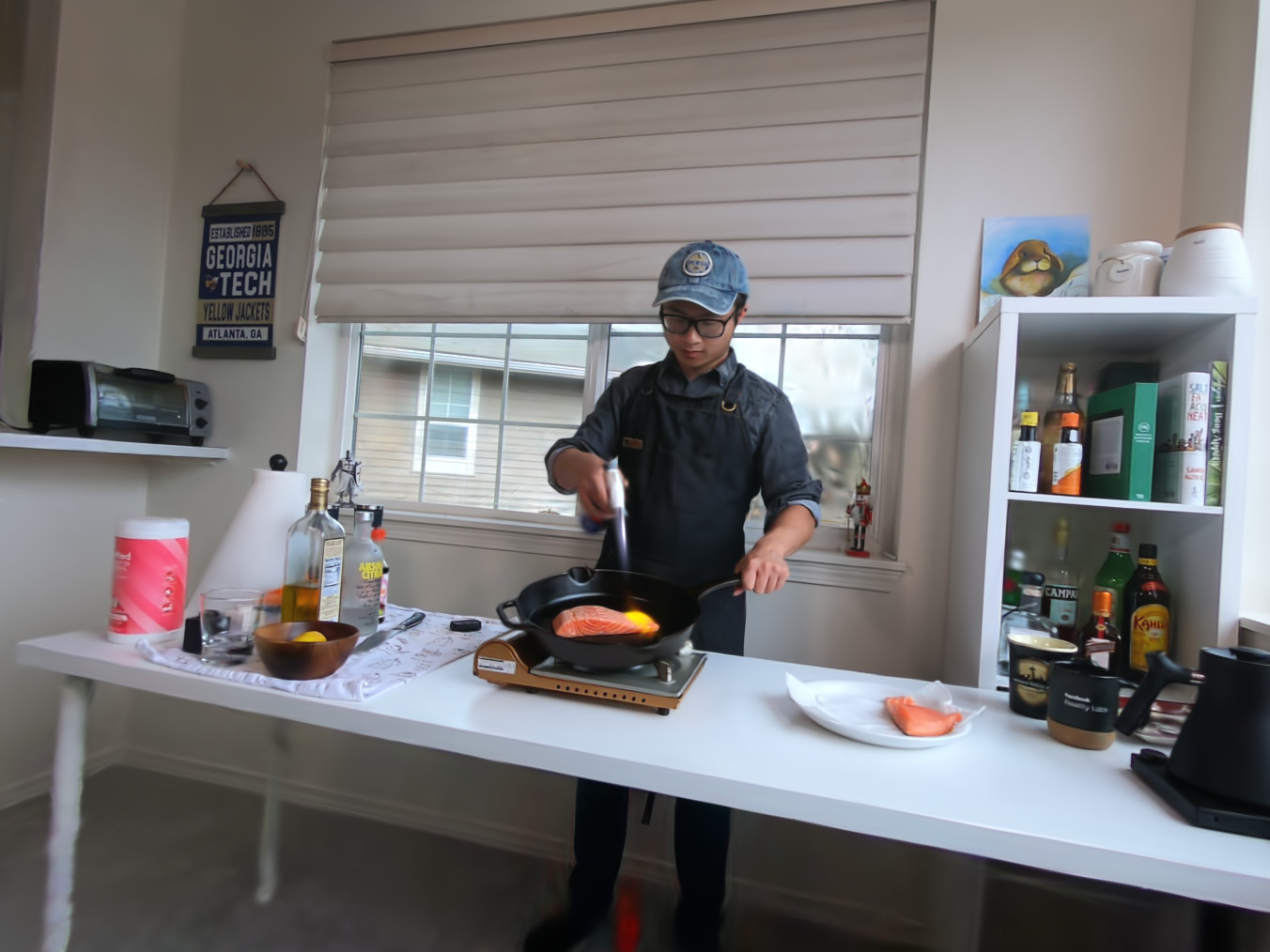} &
        \includegraphics[width=0.25\textwidth]{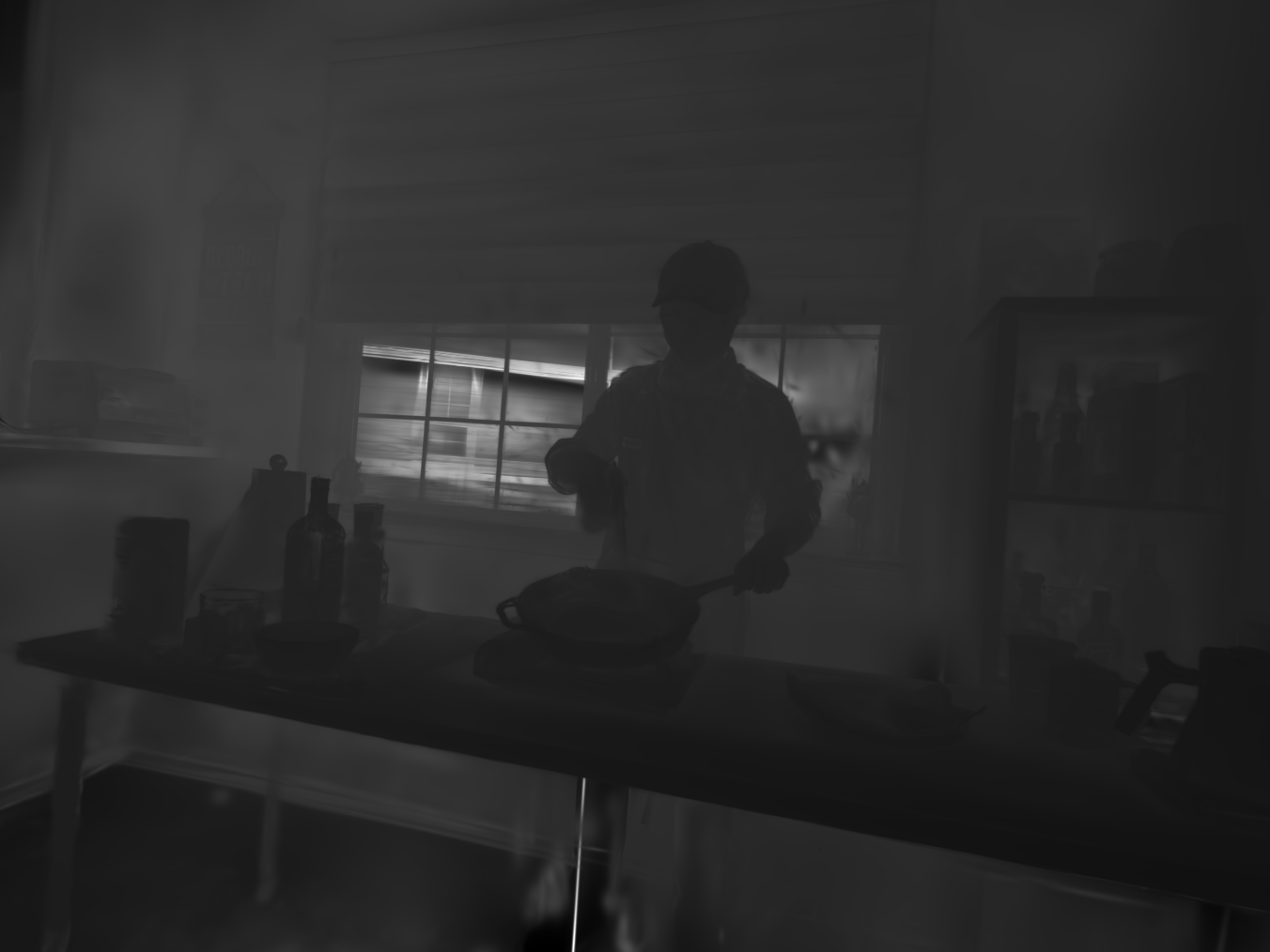}\\
        \includegraphics[width=0.25\textwidth]{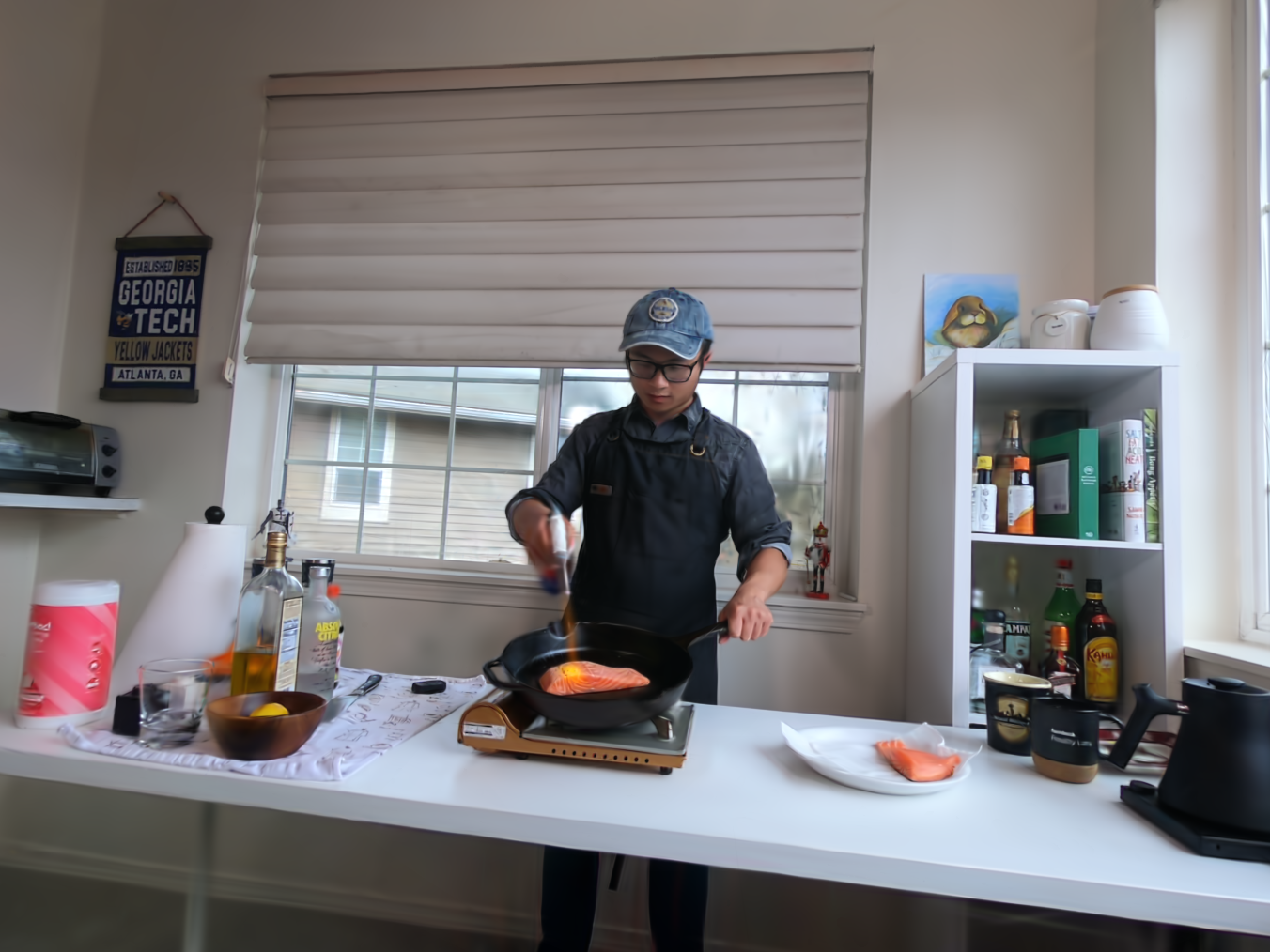} &
        \includegraphics[width=0.25\textwidth]{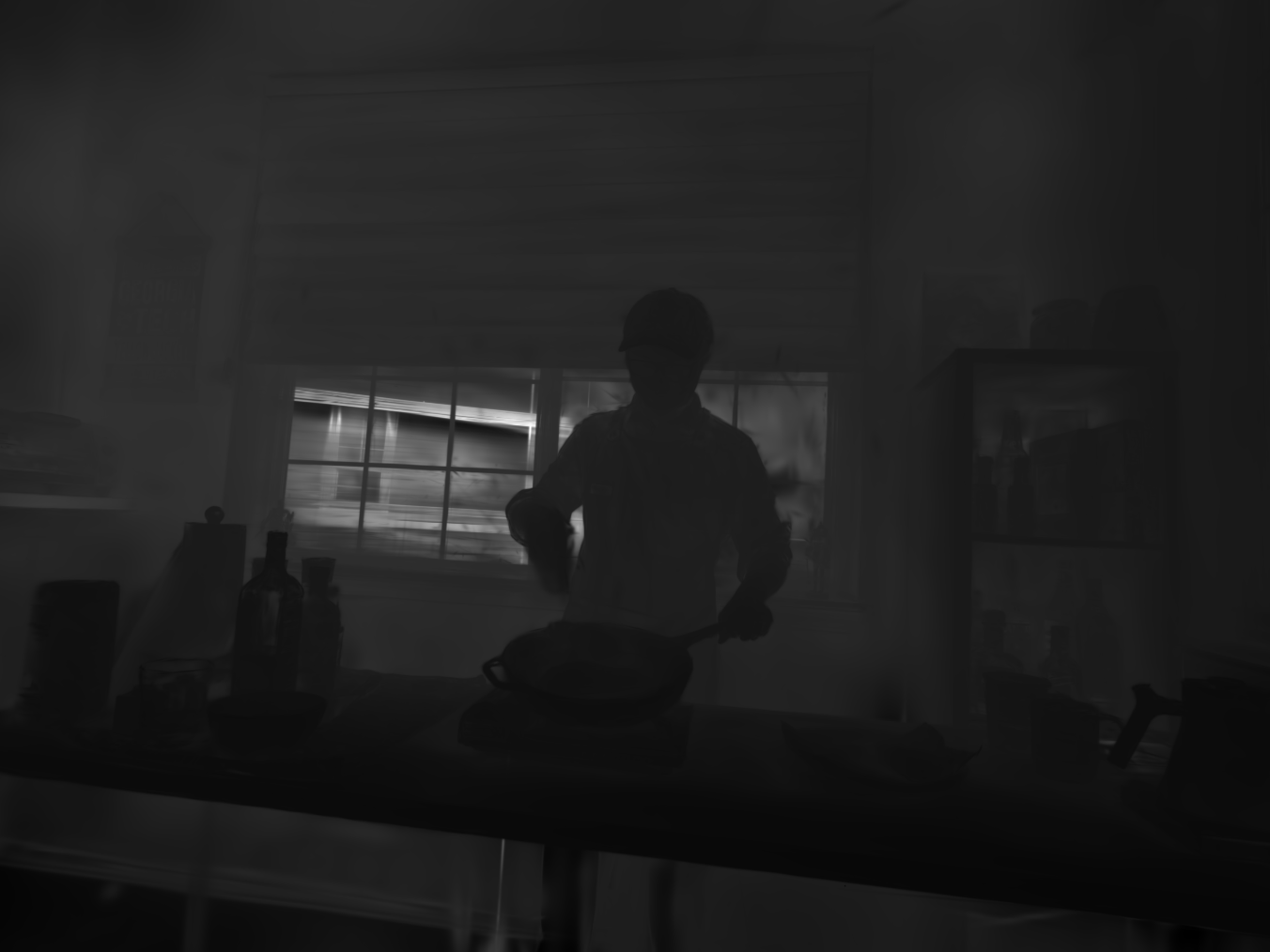} &
        \includegraphics[width=0.25\textwidth]{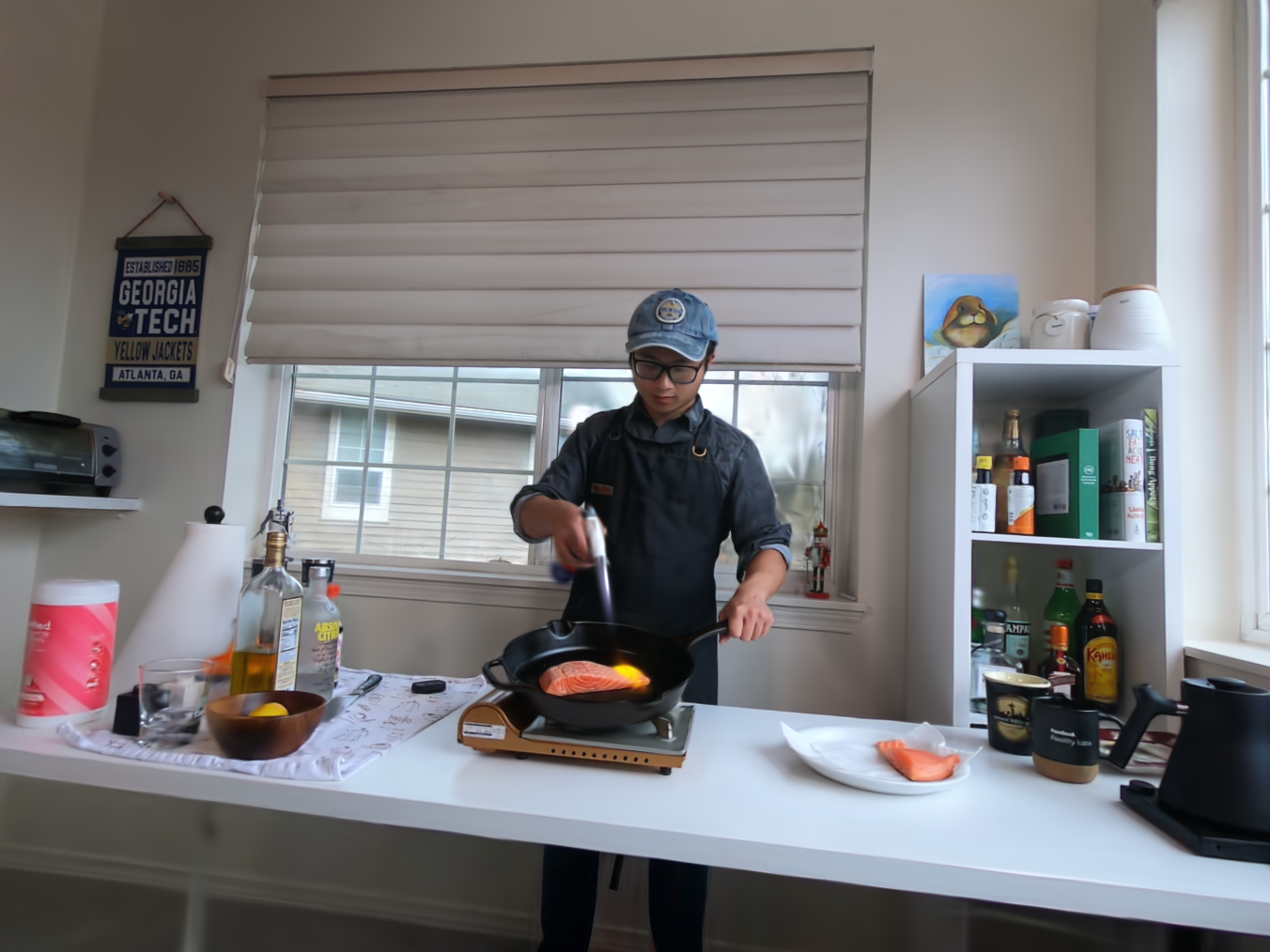} &
        \includegraphics[width=0.25\textwidth]{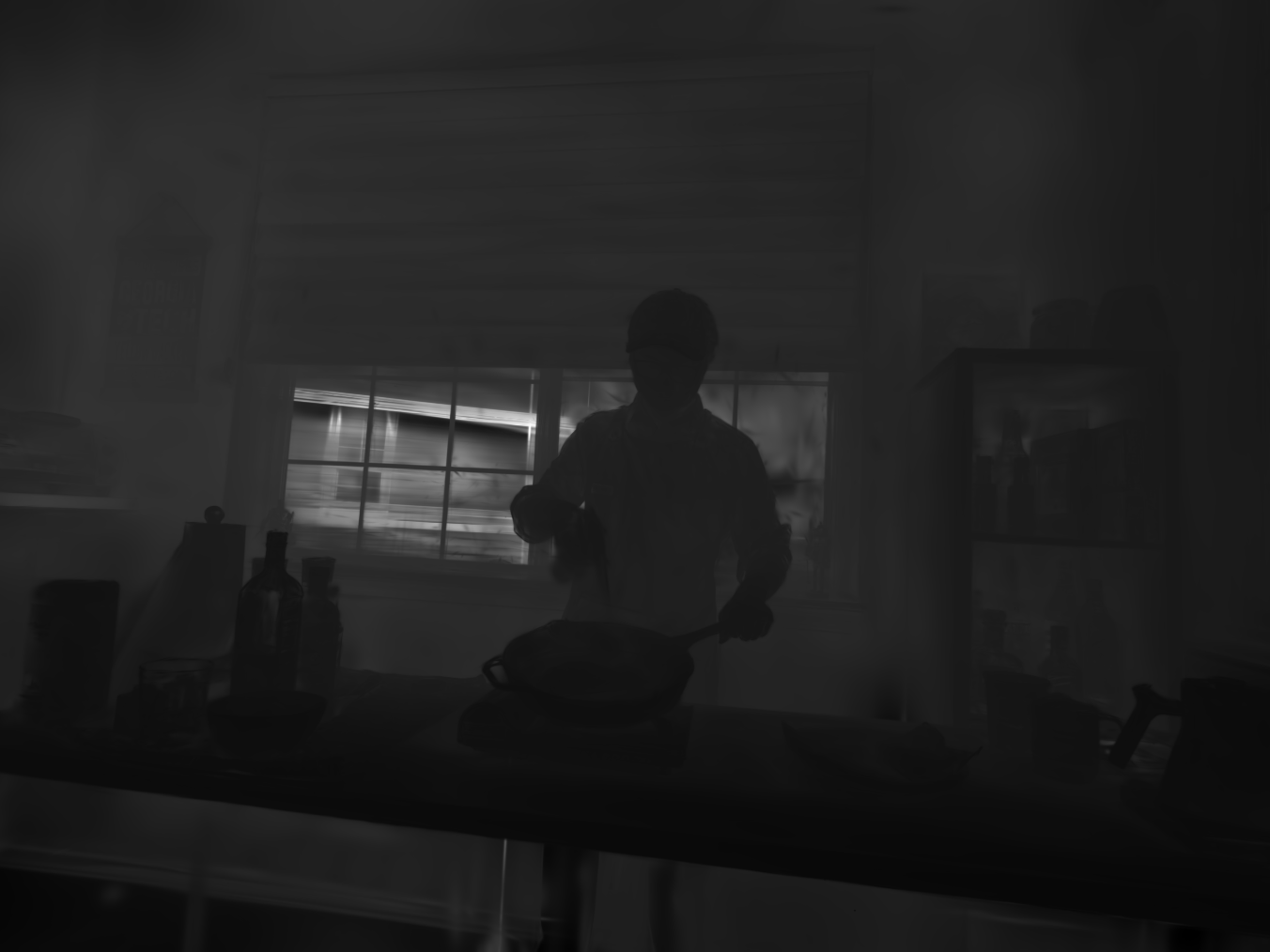}\\
        \includegraphics[width=0.25\textwidth]{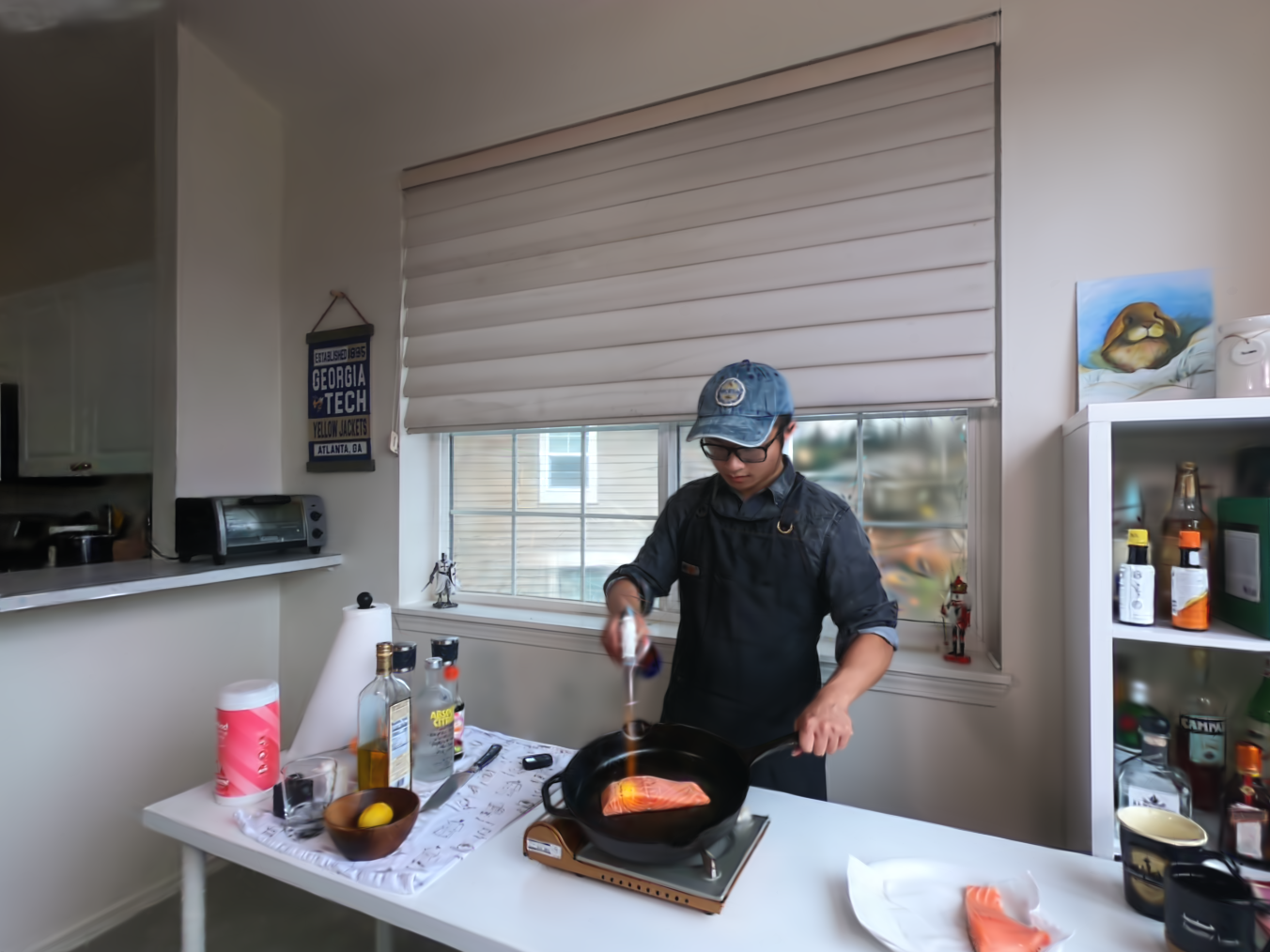} &
        \includegraphics[width=0.25\textwidth]{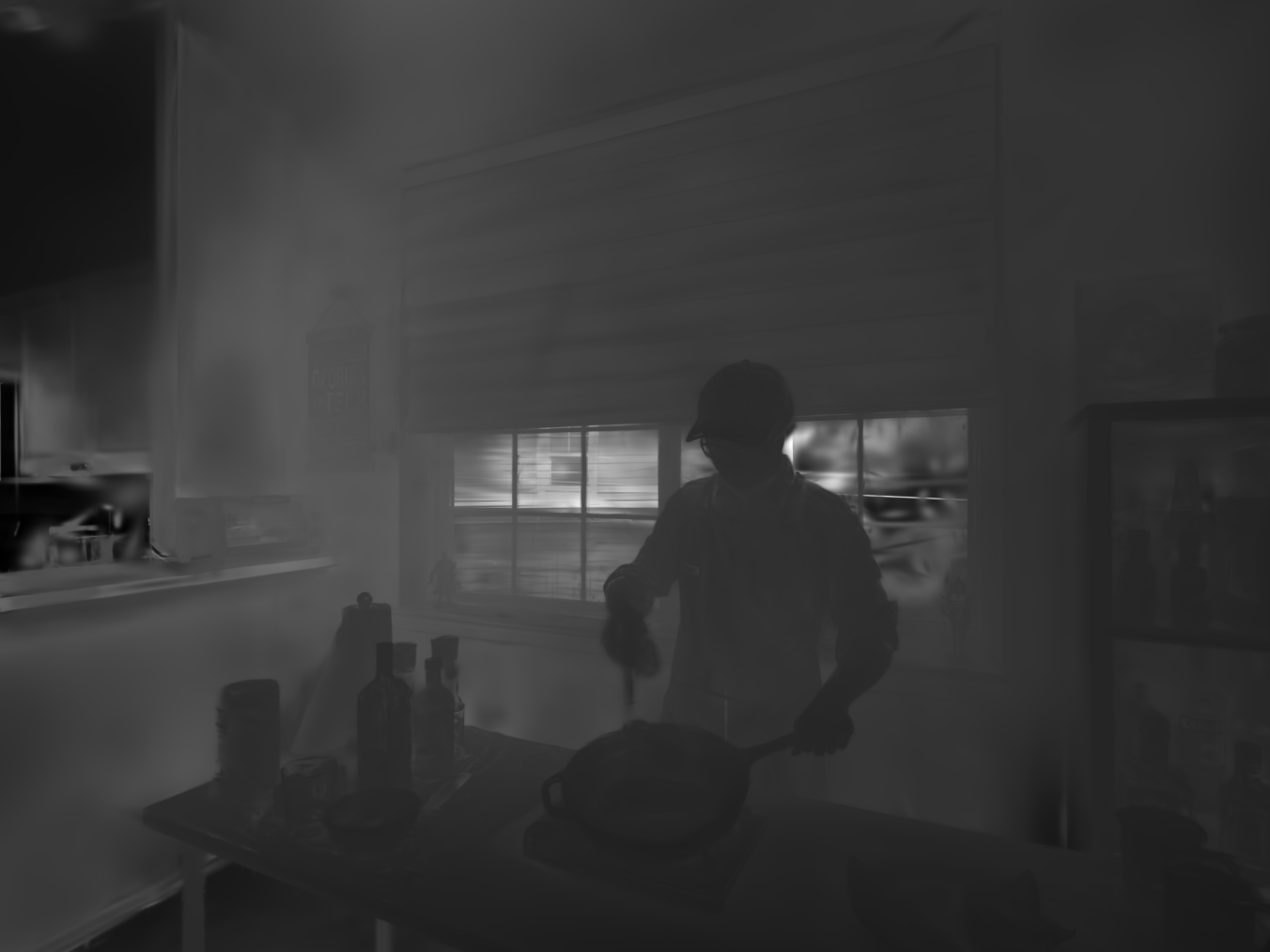} &
        \includegraphics[width=0.25\textwidth]{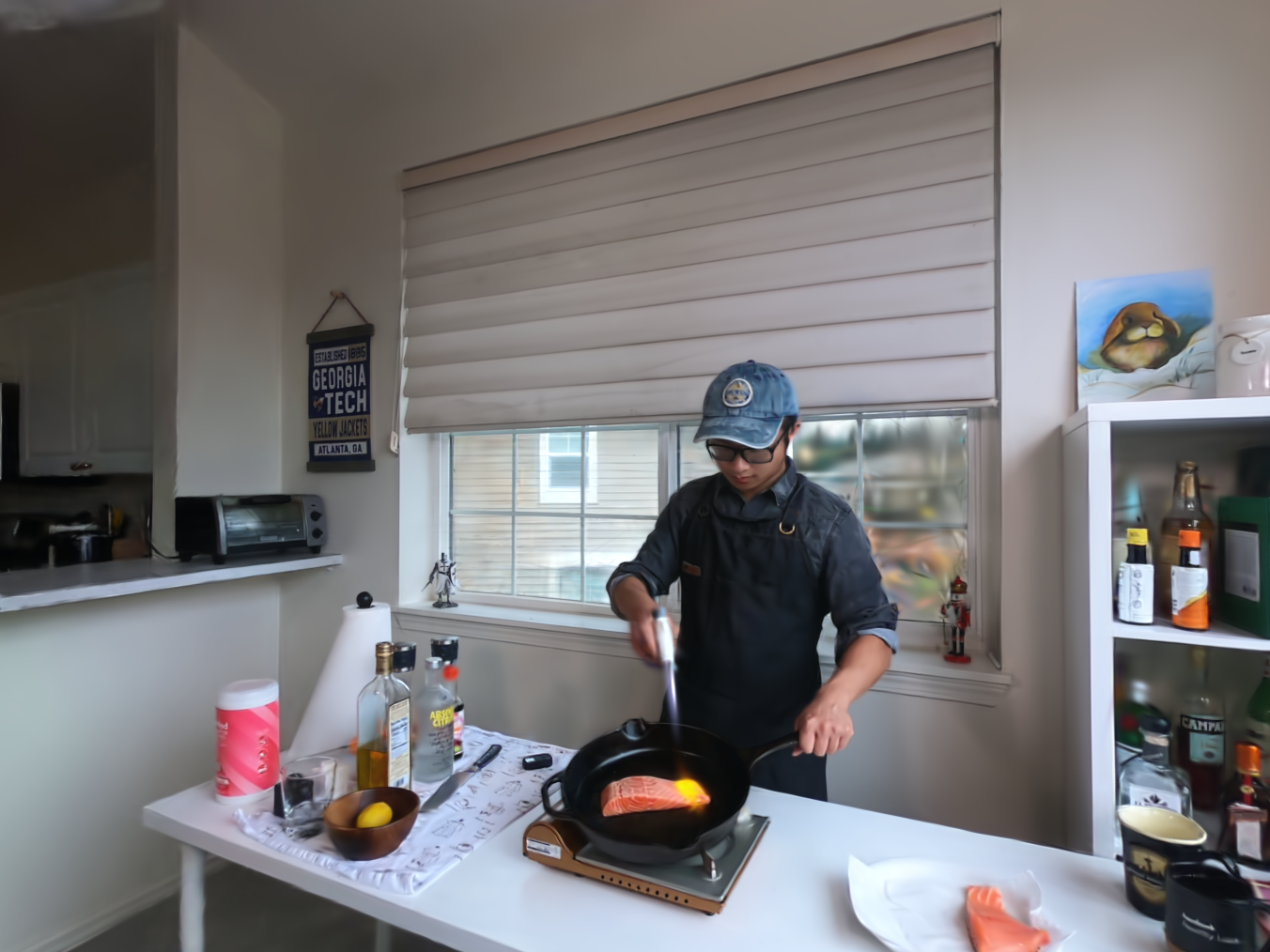} &
        \includegraphics[width=0.25\textwidth]{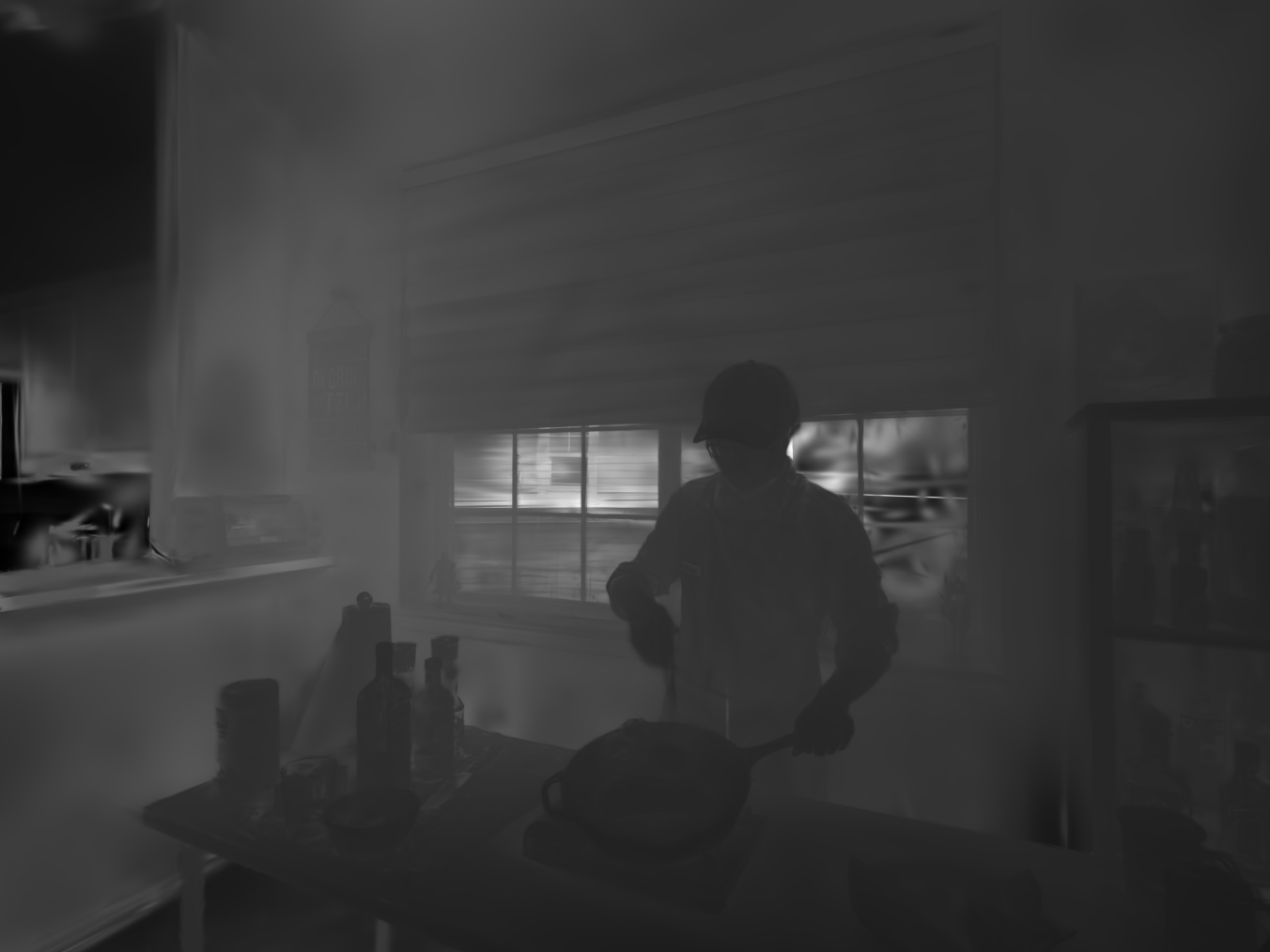}\\
        \includegraphics[width=0.25\textwidth]{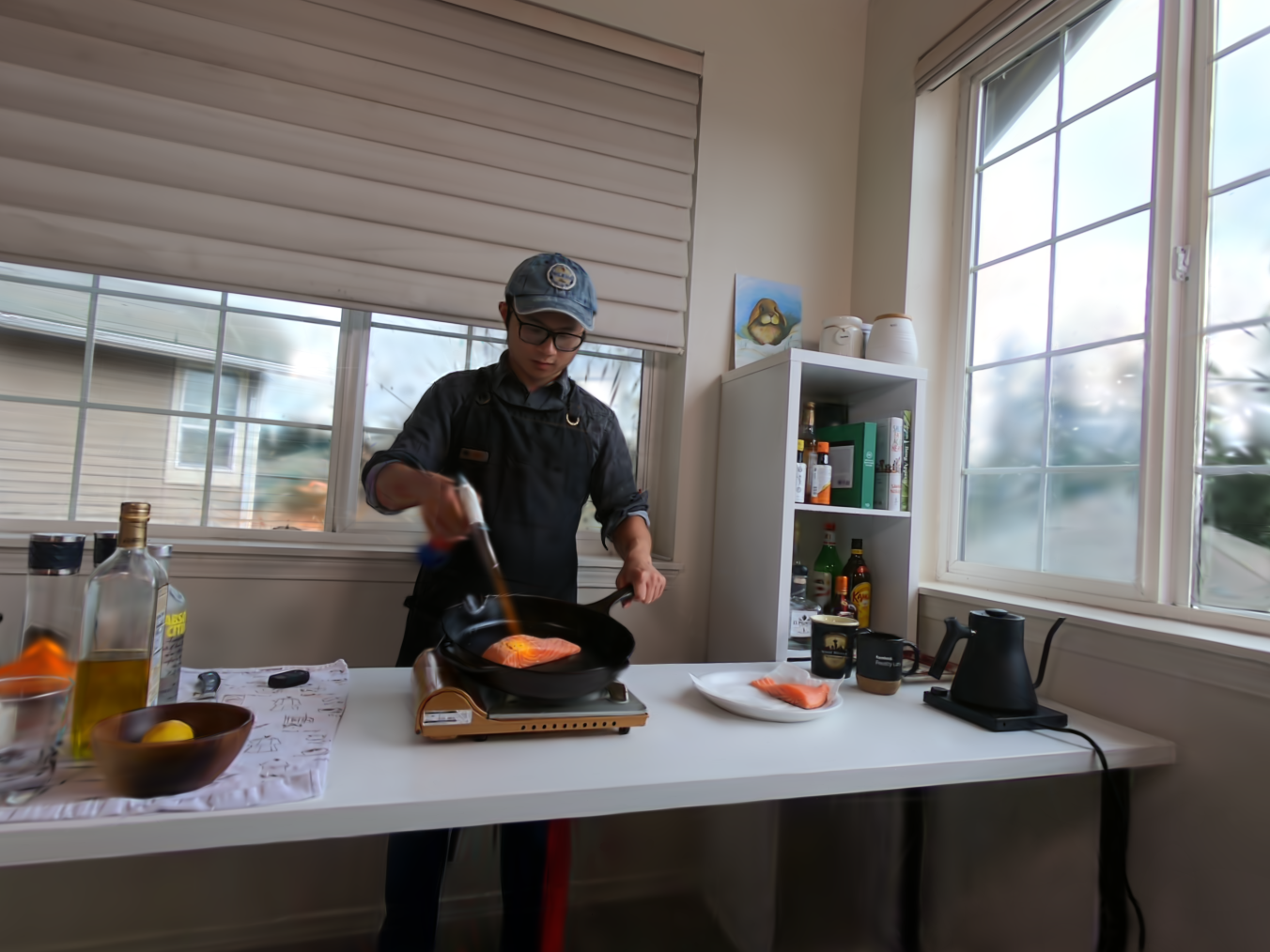} &
        \includegraphics[width=0.25\textwidth]{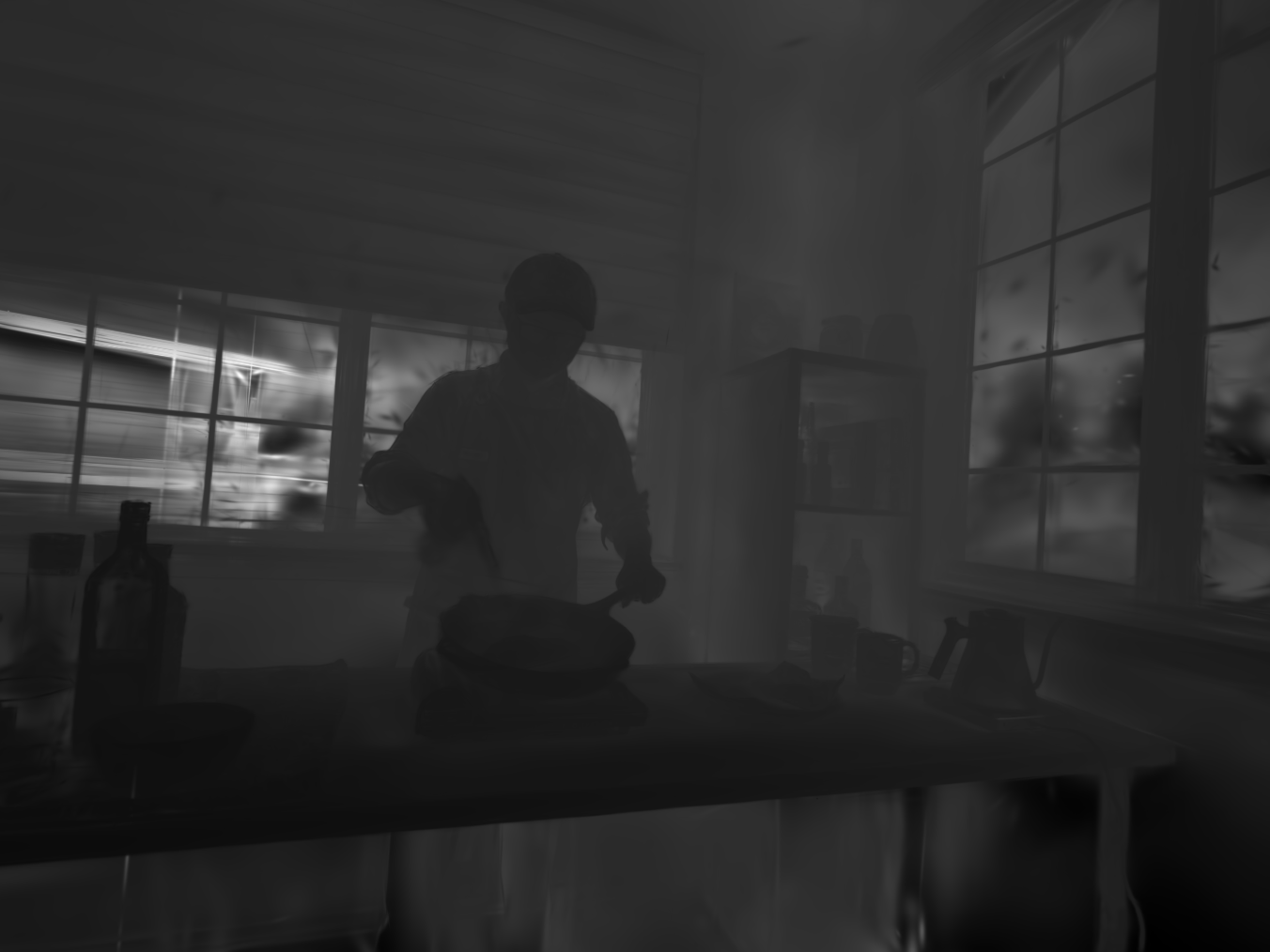} &
        \includegraphics[width=0.25\textwidth]{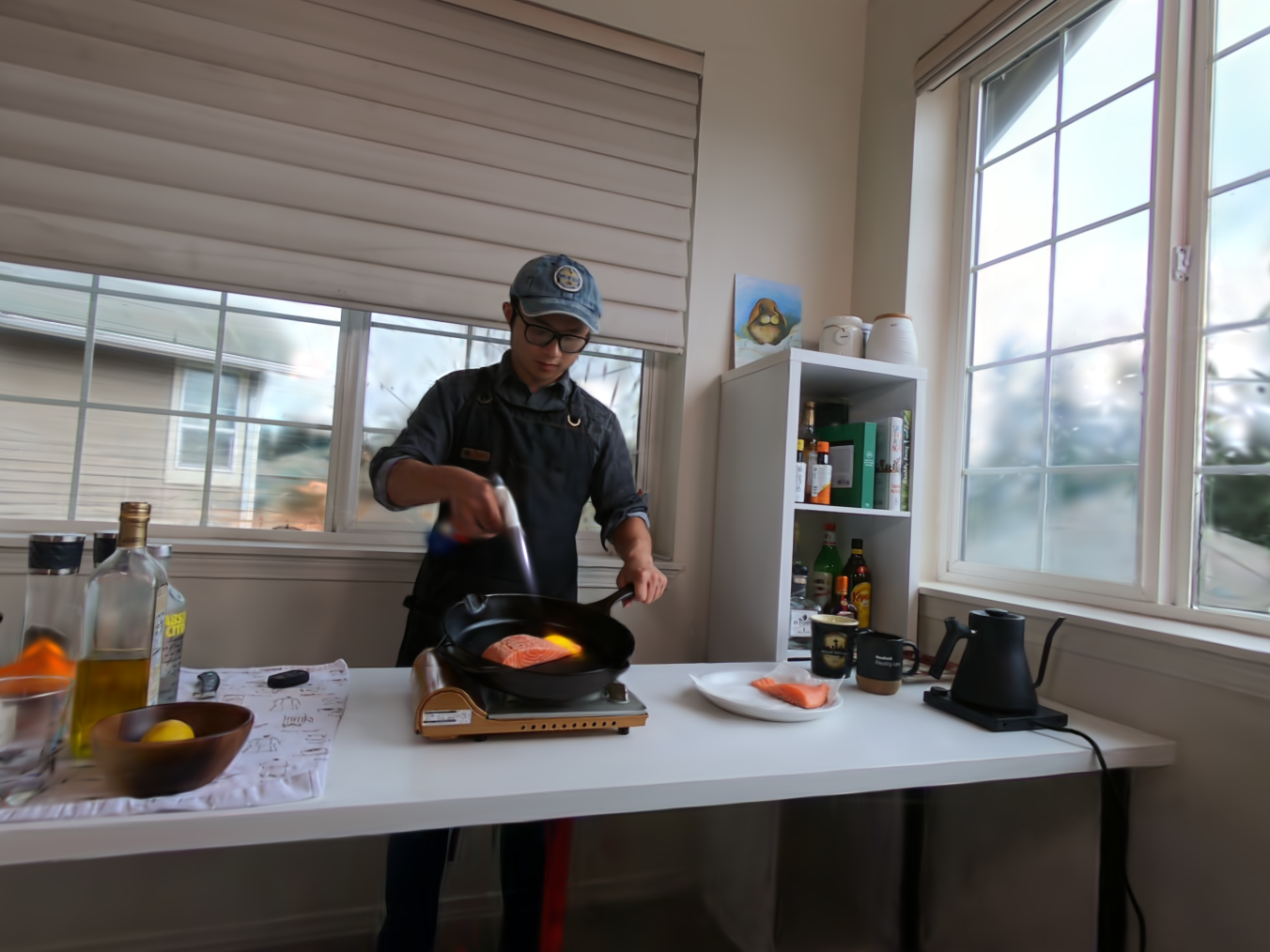} &
        \includegraphics[width=0.25\textwidth]{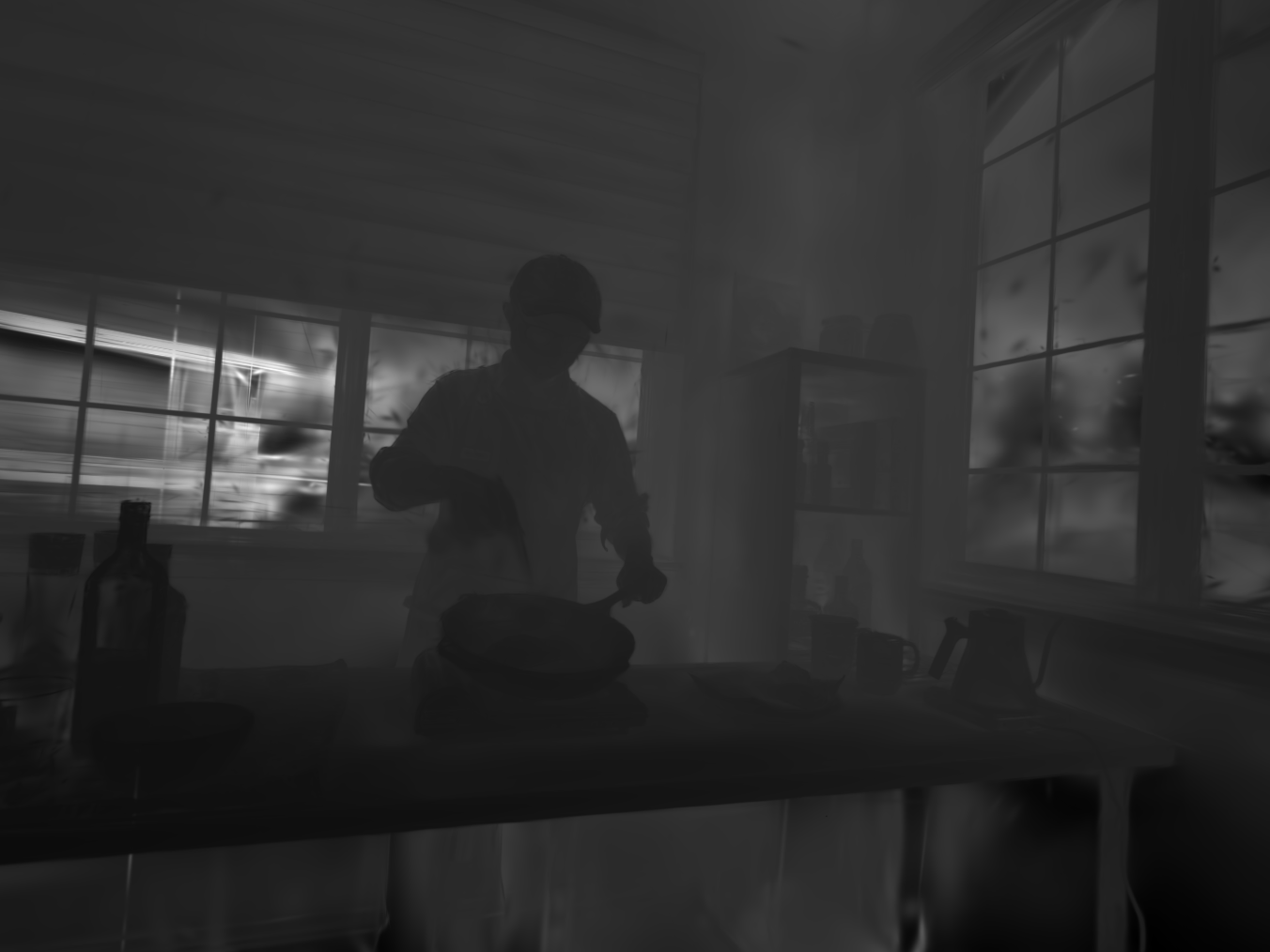}\\
        \includegraphics[width=0.25\textwidth]{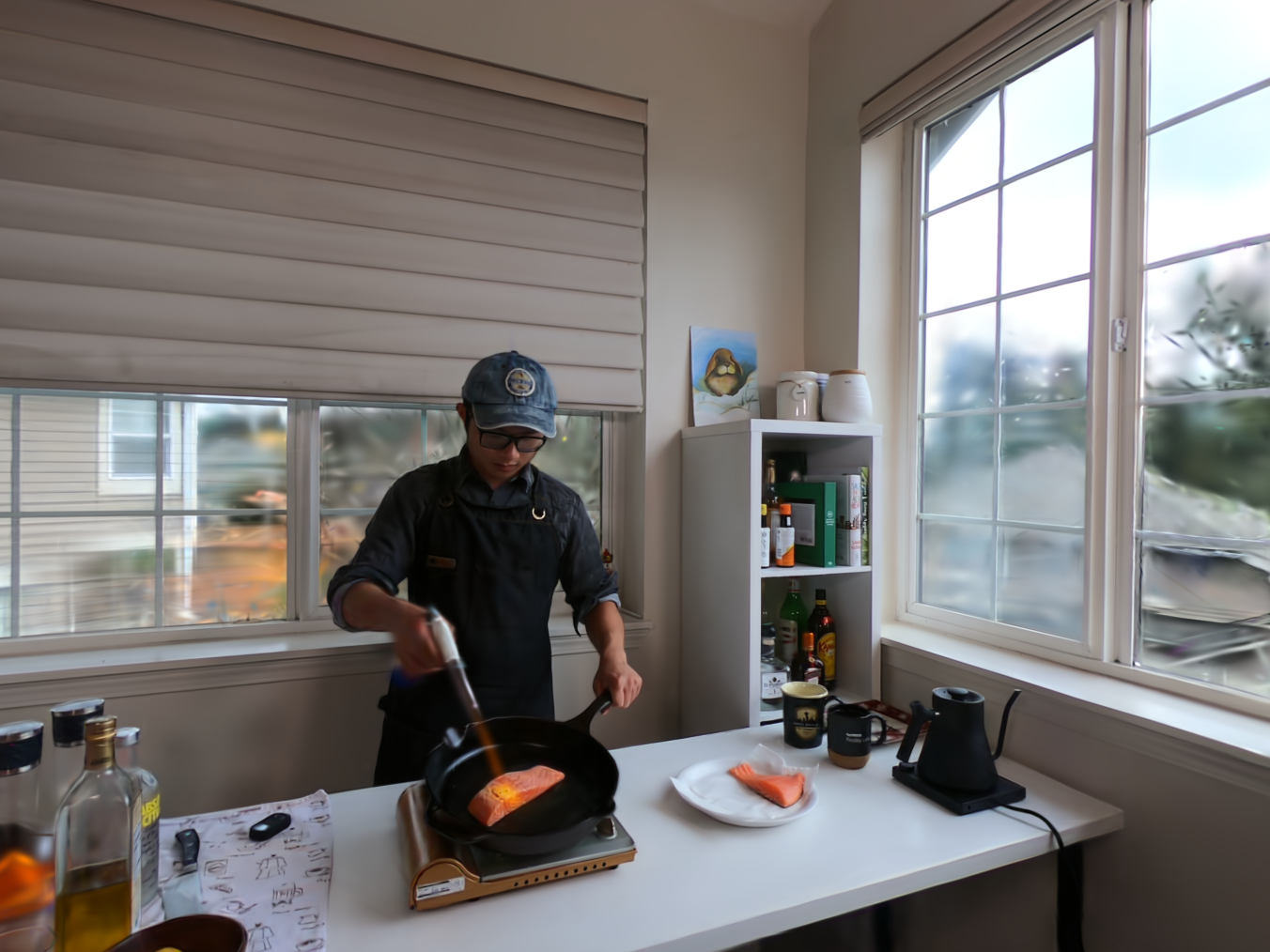} &
        \includegraphics[width=0.25\textwidth]{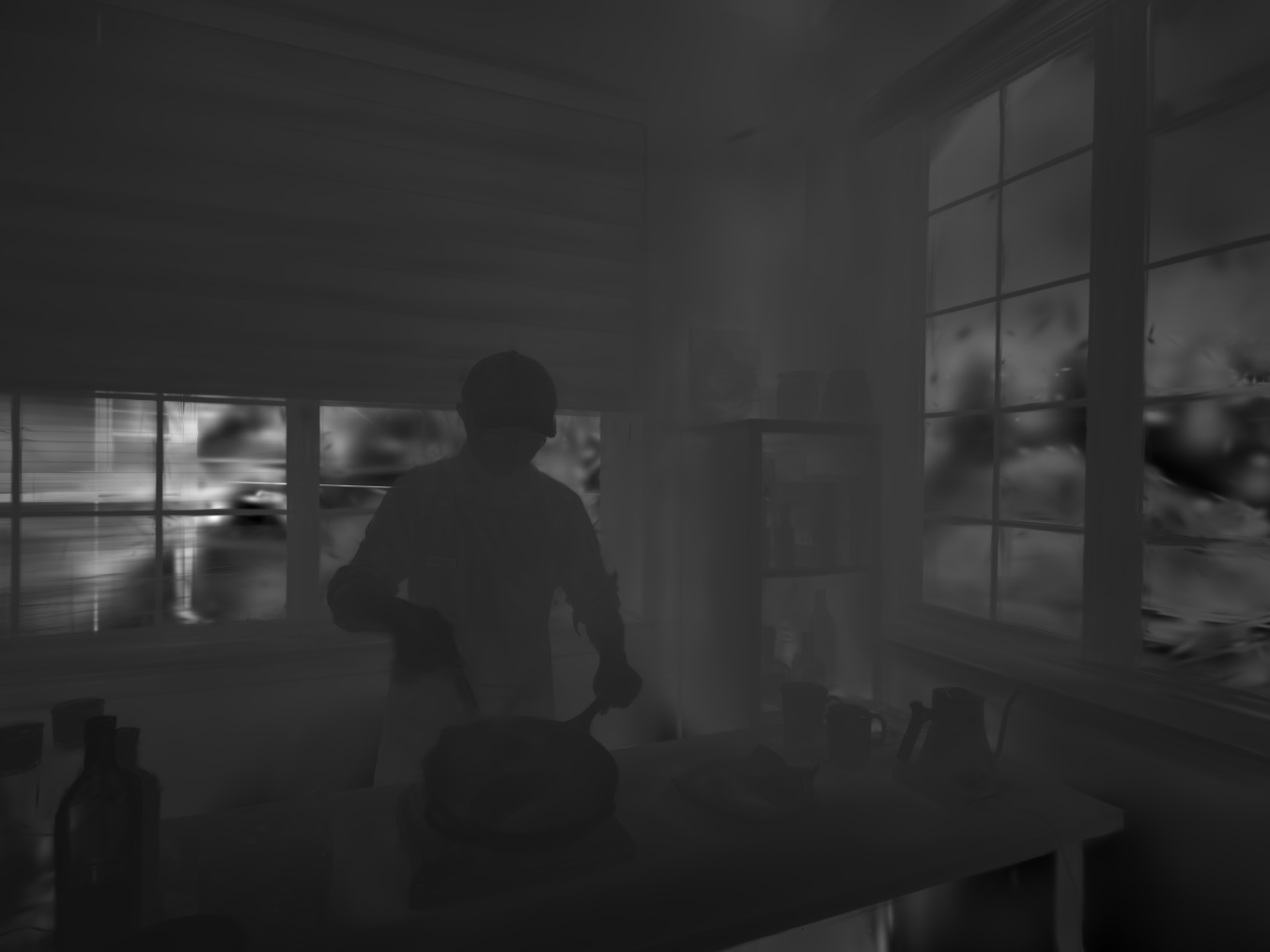} &
        \includegraphics[width=0.25\textwidth]{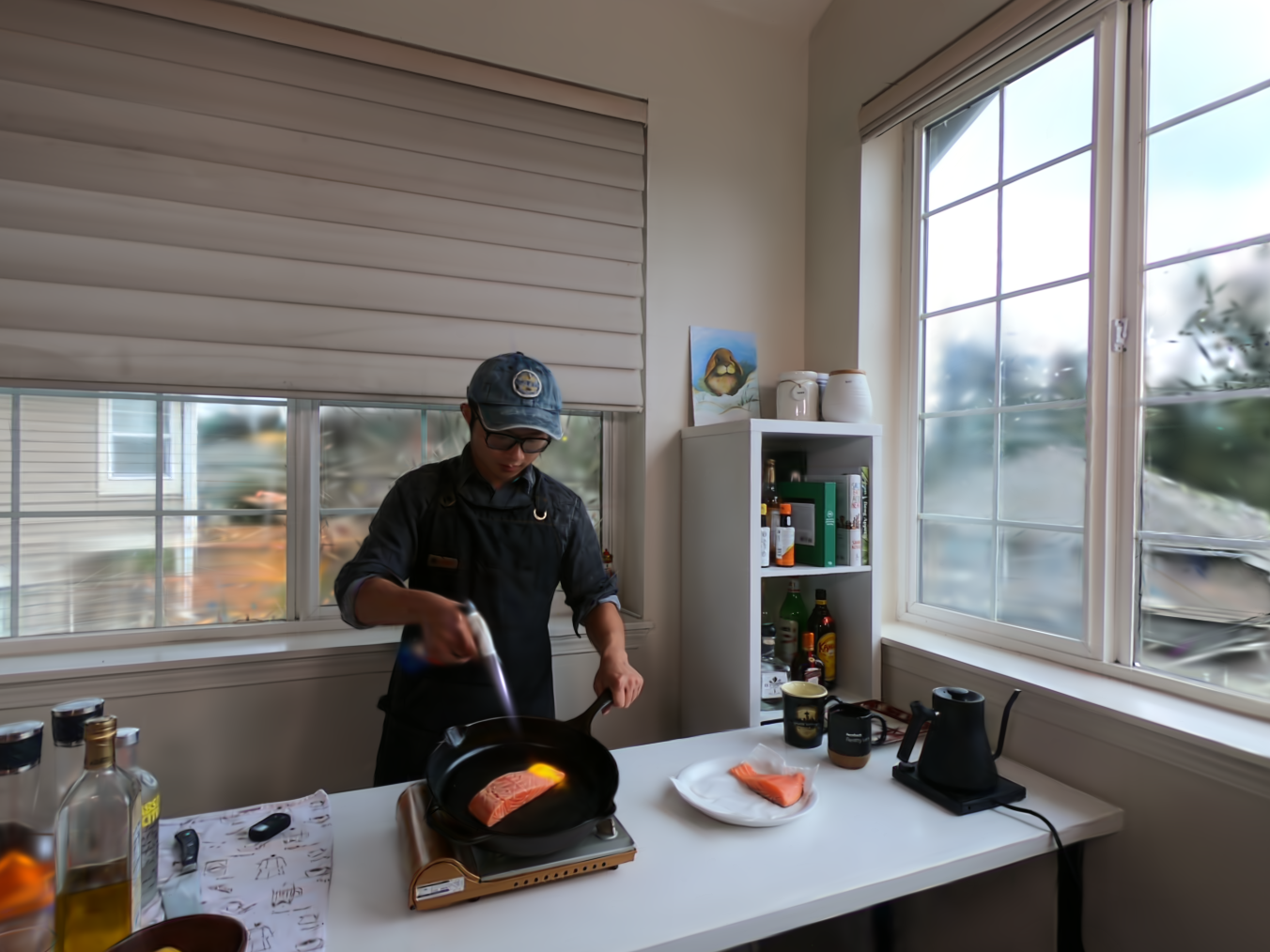} &
        \includegraphics[width=0.25\textwidth]{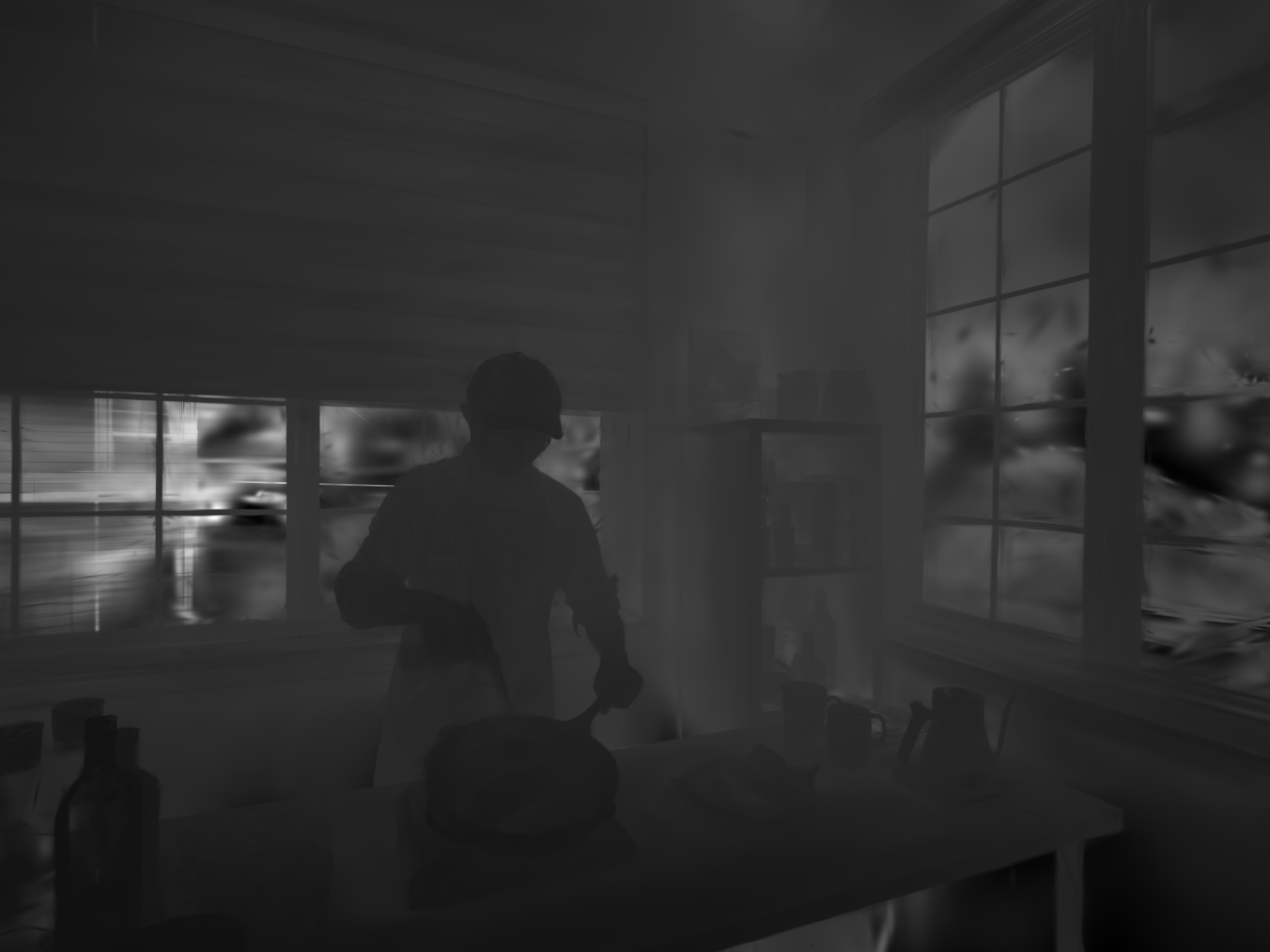}\\
    \end{tabular}
    \caption{
        \textbf{View Synthesis Results and Depths on Plenoptic Video Dataset.}
        }
    \label{sup:dynerf_depth}
    \vspace{7mm}
\end{figure*}

As shown in Figure~\ref{sup:inter_hyper}, we show more results on scene \textbf{\textit{americano}}, \textbf{\textit{chickchicken}} and \textbf{\textit{split cookie}}. As shown in Figure~\ref{sup:dynerf_depth}, we show the rendering and depth map results on Plenoptic Video dataset rendered at more viewpoints and time.

\end{document}